\documentclass{article} %
\usepackage{iclr2024_conference,times}

\usepackage{amsmath,amsfonts,bm, amsthm}

\def\1{\bm{1}}

\DeclareMathAlphabet{\mathsfit}{\encodingdefault}{\sfdefault}{m}{sl}
\SetMathAlphabet{\mathsfit}{bold}{\encodingdefault}{\sfdefault}{bx}{n}

\usepackage{hyperref}
\usepackage{url}
\usepackage{amsmath}

\title{Deep Neural Network Initialization \\ with Sparsity Inducing Activations}

\author{Ilan Price\textsuperscript{*, \textdagger}, Nicholas Daultry Ball\textsuperscript{*}, Samuel C.H.~Lam\textsuperscript{*}, Adam C.~Jones\textsuperscript{*} \& Jared Tanner\textsuperscript{*}\\
(*) Mathematical Institute, University of Oxford \\
(\textdagger) The Alan Turing Institute \\
\texttt{\{ilan.price,nicholas.daultryball,samuel.lam,adam.c.jones,tanner\}} \\ \texttt{@maths.ox.ac.uk} \\
}

\usepackage{graphicx, tabularx} %
\graphicspath{{Figures/}{./}} %

\usepackage{amsmath, amsfonts, amssymb, amsthm, bm} %
\usepackage{units} %

\usepackage{booktabs} %

\usepackage{hyperref}

\usepackage{caption}
\usepackage{subcaption}

\usepackage{natbib}

\usepackage{multirow}

\newcommand{\relutau}{\mathsf{ReLU}_{\tau}}
\newcommand{\st}{\mathsf{ST}_{\tau}}
\newcommand{\crelu}{\mathsf{CReLU}_{\tau,m}}
\newcommand{\cst}{\mathsf{CST}_{\tau,m}}

\newcommand{\R}{\mathbb{R}}
\newcommand\numberthis{\addtocounter{equation}{1}\tag{\theequation}}
\newcommand{\bracket}[1]{\left( #1 \right)}
\newcommand{\sqbracket}[1]{\left[ #1 \right]}

\newcommand{\DS}[1]{\displaystyle{#1}}

\iclrfinalcopy %
\begin{document}

\maketitle

\begin{abstract}
Inducing and leveraging sparse activations during training and inference is a promising avenue for improving the computational efficiency of deep networks, which is increasingly important as network sizes continue to grow and their application becomes more widespread.  Here we use the large width Gaussian process limit to analyze the behaviour, at random initialization, of nonlinear activations that induce sparsity in the hidden outputs.  A previously unreported form of training instability is proven for arguably two of the most natural candidates for hidden layer sparsification; those being a shifted ReLU ($\phi(x)=\max(0, x-\tau)$ for $\tau\ge 0$) and soft thresholding ($\phi(x)=0$ for $|x|\le\tau$ and $x-\text{sign}(x)\tau$ for $|x|>\tau$).  We show that this instability is overcome by clipping the nonlinear activation magnitude, at a level prescribed by the shape of the associated Gaussian process variance map. Numerical experiments verify the theory and show that the proposed magnitude clipped sparsifying activations can be trained with training and test fractional sparsity as high as 85\% while retaining close to full accuracy.
\end{abstract}

\section{Introduction}
Improving the computational efficiency of large deep neural networks is the subject of many lines of work, including network weight pruning and quantization \citep{Blalock2020}, adapting network architectures \citep{sparse_activated}, and inducing low-dimensional structure in hidden layer outputs through low-rank, sparse, or similar structures (see Sec.~\ref{sec: related}). Two of the key quantities that determine the computational efficiency of the forward pass of a deep network are the total number of floating point operations (FLOPs) involved and the total amount of memory required. These quantities are primarily influenced by the tensor products of layer weights and layer inputs. The existence of efficient techniques for storing and computing with sparse vectors and matrices makes the sparsification of DNN weights and layer inputs/outputs a promising strategy for making the forward passes of deep networks more efficient. In contrast to the extensive literature on weight pruning and quantization, this paper investigates the training of networks with nonlinear activation functions that induce highly sparse hidden layer outputs throughout both training and inference. Sparse hidden layers are especially appealing as they naturally combine with the other methods by subselecting only a portion of the weight matrix active for that input.

To induce a target sparsity level we will leverage the fact that deep networks randomly initialized with i.i.d.\ entries generate hidden layers whose entries approach Gaussian distributions with known variance as the network width increases.   For a survey of the Gaussian process behaviour of deep networks see \cite{PDLT-2022} and references therein, and for initial derivations of feedforward networks \citep{Lee_2017}, CNNs \citep{Xiao_2018}, LSTMs and GRUs \citep{Gilboa_2019}, RNNs \citep{Chen_2018}, ResNets \citep{Yang_2017} and extra features like dropout \citep{schoenholz2017deep, Huang_2019} or batch normalization \citep{Yang_2019} and pruning \citep{Hayou_2020}.  This manuscript considers random initializations of both feedforward networks, abridged here as DNNs, of width $N_l$,
\begin{align}\label{eq:feedforward}
    h^l_j = \sum_{i=1}^{N_l} W^l_{ij} x_i^{l-1} + b_j^l, \qquad x^l = \phi(h^l) \qquad \text{for } l = 1, 2, \dots , L, 
\end{align}
and one-dimensional CNNs\footnote{Higher dimensional CNNs follow similarly but with more complex notation, see \cite{Lee_2017}.} with input channel length $C_{in}^l$ and kernel width $2k+1$, 
\begin{align}\label{eq:cnn}
    h^l_j(\alpha) = \sum_{i=1}^{C_{in}^l}\sum_{\beta=-k}^k W^l_{ij}(\beta) x_i^{l-1}(\alpha+\beta) + b_j^l, \qquad x^l = \phi(h^l) \qquad \text{for } l = 1, 2, \dots , L;
\end{align}
specifically, with a focus on activation functions $\phi$ that induce a prescribed substantial fractional sparsity at initialization, which in practice is maintained throughout training and inference on unseen data. To the best of our knowledge, highly sparsifying activations have not been studied at initialization, which may well be due to --- as we will show --- the failure of DNNs and CNNs to train with the intuitive choices of sparsifying activation functions. We analytically identify the source of this failure, which is due to an instability in the variance map of successive layers' Gaussian processes. That is, for arguably the most natural sparsifying activations, the Edge-of-Chaos (EoC) initialization \cite{poole2016exponential} needed to train very deep networks happens to coincide with introducing an instability that typically results in exponential growth of the  Gaussian process variance with depth.  Here, we introduce modified activation functions, using simple magnitude clipping, which decouples EoC initialization from the Gaussian process variance, and demonstrate that this allows training of very deep networks with activation sparsity up to 85\%.

\subsection{Main Results}\label{sec: main results}
The Gaussian process model of DNNs \eqref{eq:feedforward} in the large width limit and CNNs \eqref{eq:cnn} in the large channel limit are characterized by their associated variance.  For DNNs of width $N_l$ with Gaussian weights $w_{i,j}$ drawn i.i.d.\ $\mathcal{N}(0,\sigma_w^2/N_l)$ and bias $b_j$ drawn i.i.d.\ $\mathcal{N}(0,\sigma_b^2)$ it was shown by \cite{Lee_2017} that the pre-activation outputs $h_j^l$ approach $\mathcal{N}(0,q^l)$ where $q^l$ can be computed through the `variance map' or `length map'
\begin{align}\label{eq: variance map}
    q^l = V_\phi(q^{l-1}) := \sigma_w^2 \int_\R \left(\phi(\sqrt{q^{l-1}}\, z )\right)^2 \gamma(dz) + \sigma_b^2,
\end{align}
where $\gamma(dz)$ is the standard Gaussian measure $e^{-z^2/2}/\sqrt{2\pi} \, dz$, and $q^0 = \|x^0\|^2_2 /N^0$, $q^1 = \sigma_w^2 q^0 + \sigma_b^2$. Note that the variance map $V_\phi(\cdot)$ depends on $\phi$, $\sigma_w^2$, and $\sigma_b^2$. Typically $V_\phi$ has a single nonzero fixed point, which is essential for the resulting theory\footnote{There are, however, some notable counter-examples, with ReLU having $V_{\text{ReLU}}(q)=q+\sigma_b^2$ and consequently when $\sigma_b=0$, all $q$ are fixed points, while when $\sigma_b>0$ the variance $q^l$ increases by $\sigma_b^2$.  Conversely ELU with $\sigma_b=0$ has $V_{\mathsf{ELU}}(q)<q$ for all $q$ with $q^*=0$ the only fixed point, see \cite{murray2022activation}.}.  

The Gaussian process model for CNNs \eqref{eq:cnn} in the large channel limit is somewhat more complex due to correlation within $h^l_j(\alpha)$ and $h^l_j(\alpha')$ when $|\alpha-\alpha'|<2k+1$.  \cite{Xiao_2018} proved that this Gaussian process is characterized by the matrix 
\begin{align}\label{eq: cnn_var}
\Sigma_{\alpha,\alpha'}^l = \sigma_b^2 + \frac{\sigma_w^2}{2k+1}\sum_{\beta=-k}^k \mathbb{E}  \left( \phi(h^{l-1}_j(\alpha+\beta))\phi(h^{l-1}_j(\alpha'+\beta))\right),
\end{align}
where expectation is taken over the weights and biases. The full hidden layer vector $h^l$ then approaches the Gaussian distribution $\mathcal{N}(0,\mathcal{A}\star\sigma_{\alpha,\alpha'}^l)$ where $\mathcal{A}=\frac{1}{2k+1}I_{2k+1}$ and $\star$ denotes a two-dimensional circulant cross-correlation that accounts for the $2k+1$ overlapping locations. The covariance matrix for the CNN Gaussian process follows a similar variance map as \eqref{eq: variance map},
\begin{align}\label{eq: cnn_var_map}
  \Sigma_{\alpha,\alpha'}^l = q^l (\delta_{\alpha,\alpha'} + (1-\delta_{\alpha,\alpha'}\rho^l)),
\end{align}
where $\rho^{l-1} = q_{\alpha,\alpha'}^{l-1}/\sqrt{q_{\alpha}^{l-1}q_{\alpha'}^{l-1}}$ is the correlation coefficient of the inputs at layer $l-1$ \citep{poole2016exponential}. Assuming that both $q_a^l$ and $q_b^l$ converge quickly to $q^*$ yields the following iterative correlation map
\begin{align}\label{eq: corr map}
    \rho^{l} = R_\phi(\rho^{l-1}) = \frac{1}{q^*} \left(\sigma_w^2 \iint \phi(u_1) \phi(u_2) \gamma(dz_1) \gamma(dz_2) + \sigma_b^2 \right),
\end{align}
where $u_1 = \sqrt{q^*}\,z_1$ and $u_2 = \sqrt{q^*}(\rho^{l-1} z_1 + \sqrt{1 - (\rho^{l-1})^2}\,z_2)$, which describes how the correlation between two distinct inputs (in the DNN case) or two entries at different locations with overlapping kernels  (in the CNN case) evolves from one layer to the next, after both starting with or converging to variance $q^*$.

\begin{figure}[h]
    \centering
    \includegraphics[width=\textwidth]{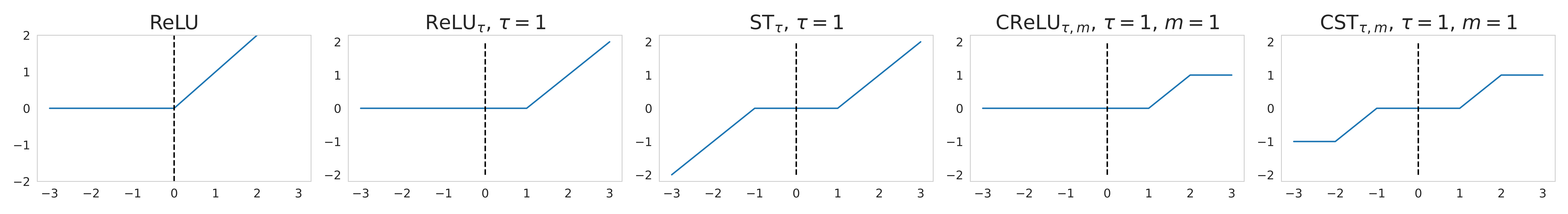}
    \caption{From left to right: ReLU (i.e. $\relutau$ with $\tau=0$), $\relutau$ ($\tau=1$), $\st$ ($\tau=1$), $\crelu$ ($\tau=1, m=1$), $\cst$ ($\tau=1, m=1$) }
    \label{fig: activation functions}
\end{figure}

Equipped with an accurate model of the hidden layer entries it is straightforward to compute the fractional sparsity that a nonlinear activation will induce, and moreover to design activation functions $\phi(\cdot)$ which induce a prescribed sparsity level in the activations. Perhaps the most obvious candidate for such sparsifying activation functions is a shifted ReLU, denoted here as $\relutau$, and defined as 
\begin{align}\label{eq:sfatrelu}
\relutau(x) = \begin{cases}
			x - \tau, & \text{if } x>\tau \\
            0, & \text{otherwise},
		 \end{cases}
\end{align}
where $\tau>0$ allows greater sparsity than the standard ReLU when $\tau=0$.\footnote{This is a continuous variant of the Forced Activation Threshold ReLU considered in \cite{kurtz2020inducing}.}  The other especially natural sparsifying activation is the SoftThreshold function $\st$ defined as
\begin{align}\label{eq:softthresholding}
\st(x) = \begin{cases}
			x - \text{sign}(x)\tau, & \text{if } |x|>\tau \\
            0, & \text{otherwise},
		 \end{cases}
\end{align}
which is an odd-function variant of the $\relutau$; see Figure \ref{fig: activation functions}. The appeal of $\st$ to induce sparsity is that it is an optimal denoiser \citep{denoise} for fixed values with additive Gaussian noise and is widely used in the compressed sensing community \citep{foucart_rauhut} to encourage sparsity.

Unfortunately, \eqref{eq:sfatrelu} and \eqref{eq:softthresholding} both suffer from a previously unreported instability of their hidden layer variance map \eqref{eq: variance map} which impedes training DNNs and CNNs with these activation functions. Specifically, both \eqref{eq:sfatrelu} and \eqref{eq:softthresholding} have the property that when initialized on the EoC, their associated $V_{\phi}(q)$ has a single fixed point $q^*$ with $V'_{\phi}(q^*)=1$, and moreover $V''_{\phi}(q^*)>0$, causing $q^*$ to be only stable from the left, see Sec.\ \ref{sec: background} and \ref{sec: activation plots}. Due to the natural stochasticity of $q^l$ for finite dimensional networks, even for moderate depths, $q^l$ typically obtains a value in excess of $q^*$ and then diverges exponentially.  This phenomenon is unusual as most common activation functions have a unique stable fixed point with $V'_{\phi}(q^*)<1$.  

It is possible to regain the stability of the fixed point of the Gaussian process variance map by suitable modification of the sparsity-inducing activation functions $\phi$. This can be achieved by clipping the magnitude of these activations, \eqref{eq:sfatrelu} and \eqref{eq:softthresholding}; specifically, let 
\begin{align}\label{eq:crelu}
\crelu(x) = \begin{cases}
			0, & \text{if } x < \tau \\
            x - \tau, & \text{if } \tau < x < \tau + m\\
            m, & \text{if } x > \tau + m
		 \end{cases}
\end{align}
and
\begin{align}\label{eq:cst}
\cst(x) = \begin{cases}
			0, & \text{if } |x| < \tau \\
            x - \text{sign}(x)\tau, & \text{if } \tau < |x| < \tau + m\\
            \text{sign}(x)m, & \text{if } |x| > \tau + m,
		 \end{cases}
\end{align}
as plotted in Figure \ref{fig: activation functions}.  This simple modification is sufficient to guarantee $V'_{\phi}(q^*)<1$ for $m$ bounded.  Proof-of-concept experiments in Sec.\ \ref{sec: max variant experiments} verify the efficacy of clipping to retain stability and show a trade-off with $m$ between the stability of training and the expressivity of the network.  Moreover, these experiments show that the for DNNs full test accuracy of a standard ReLU network baseline can be retained, or even slightly improved, with activation sparsity as high as 85\%.  For CNNs a similar phenomenon is observed with full accuracy of ReLU baseline being retained for 70\% sparsity, while 85\%  sparsity incurs just a 4\% accuracy drop.

\subsection{Related work}\label{sec: related}

As noted and referenced above, there is a large body of work analysing the infinite-width limit for stable initialization of very deep networks. Complementary to the sparse hidden layer results presented here is the extensive literature on weight pruning (sparsifying the weights), both after a network has been trained, as a type of network `post-processing' (see e.g.\ \cite{han2015learning}), and at initialization, such that the weights are sparse during training \citep{lee2019snip, tanaka2020pruning, hayou2020robust}. See \cite{hoefler2021sparsity} for a review of the work on weight sparsification/pruning. In contrast, sparsification of the layer outputs---often referred to as the layer `activations', or `feature maps' in the case of CNNs---has received relatively little attention. %
Those few works which do focus on sparsity in the activations tend to propose methods to increase the sparsity of an existing, pre-trained network via fine-tuning. \cite{georgiadis2019accelerating} proposes inducing sparse activations (feature maps) in already-trained CNNs by fine-tuning them with added $L_1$ regularisation, penalising the $L_1$ norms of all feature maps. \cite{kurtz2020inducing} instead fine-tune ReLU networks with Hoyer regularisation, and introduce the FATReLU (defined as $0$ if $x<\tau$ else $x$) for some threshold $\tau$ to increase the hidden layer sparsity.  %

\subsection{Organisation of the paper}\label{sec:outline}

Section \ref{sec:instability} analyses natural candidates for sparsifying activation functions \eqref{eq:sfatrelu} and \eqref{eq:softthresholding}, to show why very deep networks using these activation functions will almost always fail to train for $\tau>0$.  Section \ref{sec:clipped}, analyses the magnitude-clipped variants \eqref{eq:crelu} and \eqref{eq:cst}, and shows analytically that this circumvents the most important of the aforementioned inability of the networks to train.   Section~\ref{sec: max variant experiments} validates the theory with proof-of-concept numerical experiments training DNNs and CNNs using the proposed activation functions. Finally, Section~\ref{sec:conclusions} suggests a number of lines of future work. 

\section{The instability of $\relutau$ and $\st$ as sparsifying activation functions}\label{sec:instability}

The instability of $\relutau$ and $\st$ for $\tau>0$ is a consequence of the shape of their associated variance maps $V_{\phi}(q)$, \eqref{eq: variance map} when initialized on the EoC. EoC initialization requires that the slope of the correlation map $R_{\phi}(\rho)$ at $\rho=1$ 
\begin{align}\label{eq: chi1}
    \chi_{1,\phi} := R_\phi'(1) = \sigma_w^2 \int_\mathbb{R} (\phi'(\sqrt{q^*} z ))^2 \gamma(dz),
\end{align}
be equal to 1. $\chi_{1,\phi}$ determines the stability of the network to small perturbations of the inputs.  When $\chi_{1,\phi}<1$ the network maps perturbations together at an exponential rate (this is referred to as the ordered regime
\footnote{The ordered regime $\chi_{1,\phi}<1$ is stable to perturbations, but for practical training is excessively so, with exponential convergence of all inputs to a single point.}), while when $\chi>1$ the perturbations diverge at an exponential rate (the chaotic regime). Having $\chi_{1,\phi}=1$ but $R(\rho)>\rho$ for $\rho$ near 1 results in a stable convergence of correlations at a sub-exponential rate, and also helps to prevent exploding or vanishing gradients; for details see \cite{Yang_2017} and App.\ \ref{sec: background}. As noted in Section \ref{sec: main results} preserving $\chi_{1,\phi}=1$ depends on $q^l$ converging to $q^*$, and thus also requires that the variance map $V_{\phi}(q)$ be sufficiently stable around $q^*$. Unfortunately it is impossible for $\relutau$ or $\st$ to meet both of these requirements.

\begin{figure}[t]
    \centering
    \includegraphics[width=\textwidth]{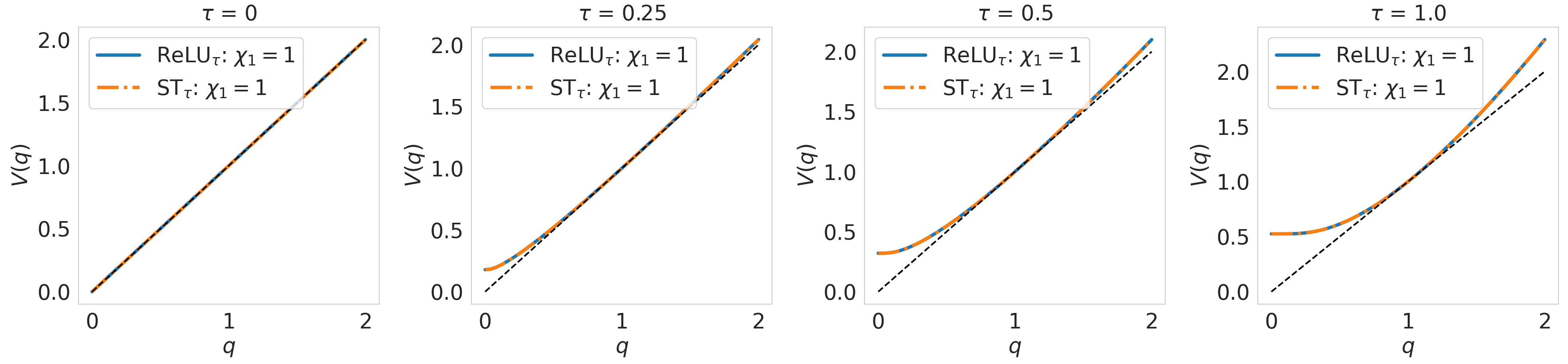}
    \caption{Variance maps for $\relutau$ and $\st$ with $(\sigma_w,\sigma_b)$ on the EoC, for different values of $\tau$. Here $q^*=1$ is used to compute $\chi_{1,\relutau}$. The dashed line is the identity map. The curves for $\relutau$ and $\st$ overlap exactly for a given $\tau$. Note, however, that a fixed value of $\tau$ corresponds to substantially different activation sparsities for $\relutau$ and $\st$. Figure \ref{fig:variance map sparsity} in App. \ref{sec: non max variants experiments} compares $V(q)$ for fixed output sparsities.}
    \label{fig:sfatrelu and st variance map}
\end{figure}

The variance maps $V_{\phi}(q)$ for $\phi = \relutau$ and $\st$ are computed in App.~\ref{sec: activation plots}, tabulated in Table~\ref{tab:instability_equations}, and plotted in Figure \ref{fig:sfatrelu and st variance map} with $(\sigma_w,\sigma_b)$ on the EoC for different values of $\tau$. The key observation is that for both $\relutau$ and $\st$, the slope $V'_{\phi}(q)$ at the fixed point $V_{\phi}(q^*)=q^*$ is equal to $\chi_{1,\phi}$, and thus we necessarily have that $V'_{\phi}(q^*)=1$ on the EoC. 

When $\tau=0$, $\relutau$ is just the standard ReLU function, and $\st$ collapses to the identity. In both cases, when $(\sigma_w,\sigma_b)$ lie on the EoC, $V'_{\relutau}(q) = 1$ for all $q$, making any $q$ a fixed point of $V_{\relutau}$; see the left-most plot in Figure \ref{fig:sfatrelu and st variance map}.

For $\tau>0$ however, the second derivative $V''_\phi(q^*)$ is strictly positive, causing networks with $\relutau$ and $\st$ with $\tau>0$ to have unstable Gaussian process variance propagation and consequently prove effectively impossible to train for large sparsity. We verify this claim experimentally -- see App.\ \ref{sec: non max variants experiments} for the experimental details and Table \ref{tab: results sfatrelu_max and st_max} for the results.

\begin{table}[]
    \centering
\begin{center}
\begin{tabular}{ |c|c|c|c|c|c|c| } 
 \hline
 $\phi$&$V’_\phi(q)$& $\chi_{1,\phi}$ at $q=q^*$ & $V''_\phi(q)$ & $\hat{\tau}_\phi(s,q^*) $ & $\sigma_w^2$ \\ \hline
 $\relutau$&$\sigma_w^2 \Phi\left(-\frac{\tau}{\sqrt{q}}\right)$ & $\sigma_w^2 \Phi\left(-\frac{\tau}{\sqrt{q^*}}\right)$& $\Phi\left(-\frac{\tau}{\sqrt{q}}\right)^{-1} \frac{\tau e^{-\frac{\tau^2}{2q}}}{2\sqrt{2\pi}q^\frac{3}{2}}$&$\sqrt{q^*} \Phi^{-1}(s)$&$\frac{1}{1-s}$ \\ 
 $\st$&$2 \sigma_w^2 \Phi\left(-\frac{\tau}{\sqrt{q}}\right)$& $2 \sigma_w^2 \Phi\left(-\frac{\tau}{\sqrt{q^*}}\right)$ & $2\Phi\left(-\frac{\tau}{\sqrt{q}}\right)^{-1} \frac{\tau e^{-\frac{\tau^2}{2q}}}{2\sqrt{2\pi}q^\frac{3}{2}}$
 &$\sqrt{2q^*} \text{erf}^{-1}(s)$ & $\frac{1}{1-s}$\\ 
 \hline
\end{tabular}
\end{center}
\caption{Variance and correlation map derivatives for $\relutau$ and $\st$ where $\Phi(\cdot)$ is the CDF of a standard normal distribution.  The correspondence of $V'_{\phi}$ and $\chi_{1,\phi}$ means that EoC initialization implies that $V'_{\phi}(q^*)=1$ for all $\tau>0$.  Moreover, the second derivative of the variance map $V''_\phi$ is strictly positive for every $q$, resulting in instability of the fixed point $V_{\phi}(q^*)=q^*$.  The simplified expression of $\sigma_w^2$ in the final column occurs when $\chi_{1,\phi}=1$.}
\label{tab:instability_equations}
\end{table}

In practice, we choose $\tau$ to control the `expected sparsity' $s$ (proportion of zeros) after applying different activation functions. Assuming the activation function is applied to a random Gaussian vector with independent entries, each with variance $q^*$, we have  %
\begin{equation}
s = \mathbb{P}(\phi(Z) = 0), \quad Z \sim \mathcal{N}(0,q^*).
\end{equation}
The optimal (smallest) $\tau$ for a DNN with activation function $\phi = \relutau$ or $\st$ and activation sparsity $s$, denoted as $\hat{\tau}_\phi(s,q^*)$, is computed in App.\ \ref{sec: clipped jacobian} and tabulated in Table \ref{tab:instability_equations}.  Substituting $\hat{\tau}_\phi(s, q^*)$ into the expression for $\chi_{1,\phi}$ in Table~\ref{tab:instability_equations} results in the value of $\sigma_w^2$ on the  EoC, which turns out to be independent of $q^*$ for both $\relutau$ and $\st$, and which thus makes achieving approximate dynamical isometry \citep{pennington2018emergence} with these networks impossible, see App \ref{sec: background}.

\section{Gaining stability and trainability via clipping}\label{sec:clipped}

Clipping the maximum output magnitude of $\relutau$ and $\st$ allows the EoC condition $\chi_{1,\phi}=1$ to be satisfied while ensuring $V'_{\phi}(q^*)<1$. The clipped activation functions are denoted as $\crelu$ and $\cst$ in \eqref{eq:crelu} and \eqref{eq:cst} respectively. This allows $\crelu$ and $\cst$ to generate hidden layer outputs which are both very sparse and trainable when initialized at the EoC. 

The associated variance and correlation maps for $\crelu$ and $\cst$ are computed and plotted for a variety of sparsity factors $\tau$ and clipping levels $m$ in App.\ \ref{sec: clipped plots}. %
The stability of the fixed point $q^*$ when $\chi_{1,\phi} = 1$ follows from the  expressions for the gradient of their variance map $V'_{\phi}(q)$ and the slope $\chi_{1,\phi}$:
\begin{align}
    V'_{\crelu}(q)= \sigma_w^2 \left( -\frac{m}{\sqrt{2 \pi q}} e^{-\frac{1}{2} \big(\frac{m+\tau}{\sqrt{q}}\big)^2}  + \frac{1}{2} \left(\text{erf}\left(\frac{m + \tau}{\sqrt{2q}}\right) - \text{erf}\left(\frac{\tau}{\sqrt{2q}}\right) \right) \right), \label{eq: vprime of clipped relu}
\end{align}
\begin{align}
V'_{\cst}(q) = 2 V'_{\cst}(q),
\end{align}
\begin{align}
    \chi_{1,\crelu} = \frac{\sigma_w^2}{2} \left( \text{erf}\left(\frac{m + \tau}{\sqrt{2q^*}}\right) - \text{erf}\left(\frac{\tau}{\sqrt{2q^*}}\right) \right),
\end{align}
\begin{align}
    \chi_{1,\cst}= 2 \chi_{1,\crelu}.
\end{align}
Crucially, %
\begin{align}
    \chi_{1,\crelu} & = V'_{\crelu}(q^*) + \frac{\sigma_w^2 m}{\sqrt{2 \pi q^*}} \exp\bracket{-\frac{(m+\tau)^2}{2q^*}} > V'_{\crelu}(q^*), \label{eg: delta v' chi sfatrelu max} \\
    \chi_{1,\cst} & = V'_{\cst}(q^*) + \frac{\sqrt{2}\,\sigma_w^2 m}{\sqrt{\pi q^*}} \exp\bracket{-\frac{(m+\tau)^2}{2q^*}} > V'_{\cst}(q^*). \label{eg: delta v' chi soft threshold max}
\end{align}
Thus, $V'_{\phi}(q^*) < 1$ at the EoC ($\chi_{\phi}=1$), making $q^*$ \textit{locally} stable for both $\phi = \crelu$ or $\cst$. The local convergence rate of $q^l$ to the fixed point $q^*$ is determined by $1 - V'_{\phi}(q^*)$. Note, however, that certain values of $m$ and $s$ induce multiple fixed points, causing $q^*$ not to be \textit{globally} stable; see Figure \ref{fig: st_max vmaps large vprime} as well as the bottom right panels of Figures \ref{fig: sfatrelu max variance map} and \ref{fig: st max variance map} in App \ref{sec: clipped plots}. 
Another effect of bounding these activation functions is that $\sigma^2_w$ at the EoC ($\chi_{1,\phi}=1$) is no longer independent of $q^*$ for a given expected sparsity $s$, although this unfortunately still does not enable us to achieve approximate dynamical isometry for these activation functions, for details see App.~\ref{sec: clipped jacobian}.
\begin{figure}[h]
    \centering
    \includegraphics[width=\textwidth,height=4cm]{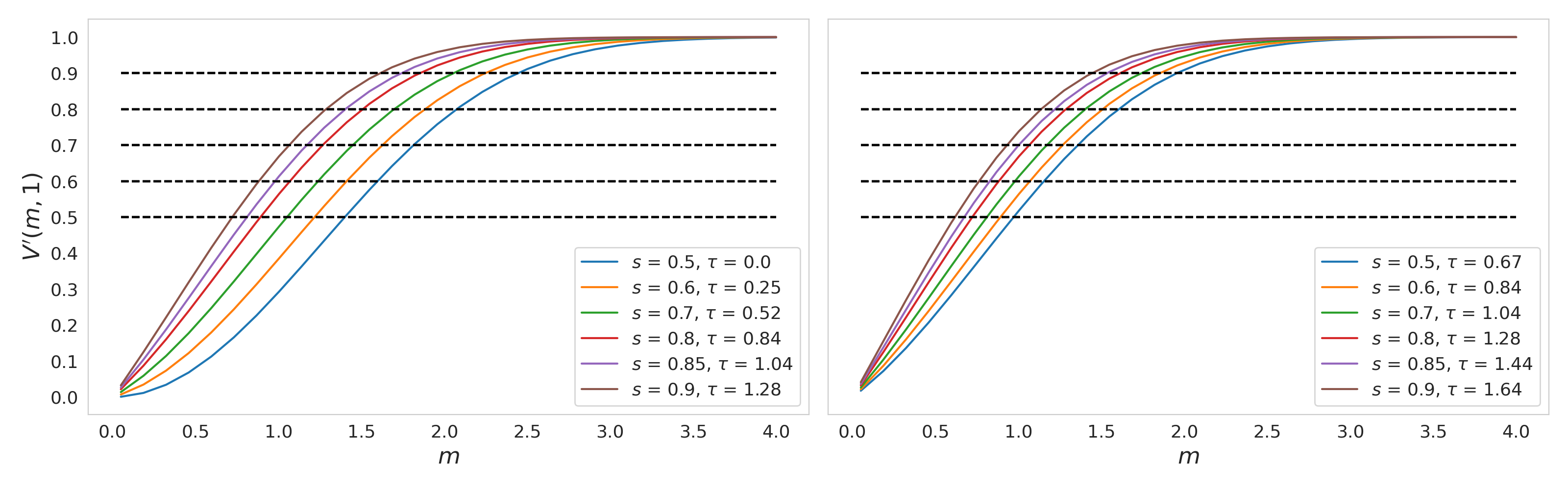}
    \includegraphics[width=\textwidth,height=4cm]{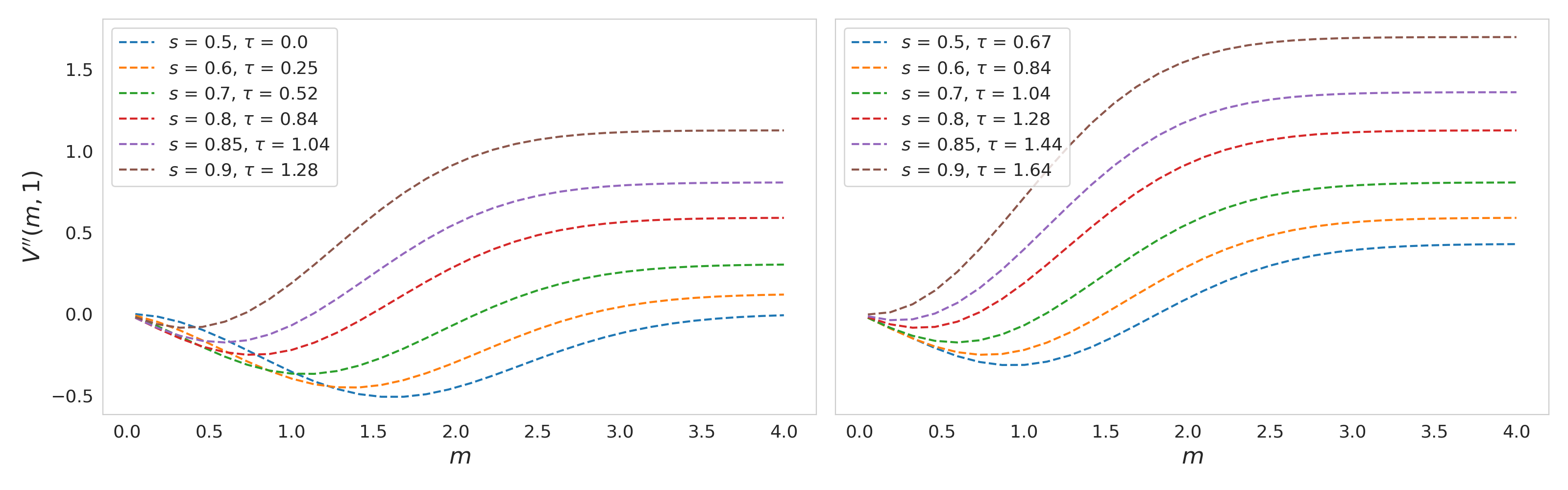}
    \caption{Plots of $V'_{\phi}(m)$ (upper) and $V''_{\phi}(m)$ (lower) and at $q^*=1$ for $\crelu$ (left) and $\cst$ (right) with for different fractional sparsities. For $V'_{\phi}(m)$ the horizontal dashed black lines are plotted at 0.5, 0.6, 0.7, 0.8, 0.9.}
    \label{fig: vprimes of m}
\end{figure}
\subsection{Choosing the clipping magnitude $m$ for $\crelu$ and $\cst$. }\label{sec: choosing m}
A larger clipping hyperparameter $m$ increases the expressivity of the network by allowing larger activation values and a wider trainable (non-zero gradient) region of its domain. It also plays a role in determining the shape of the variance map---crucially, $V'_\phi(q^*)$ (see Equation \eqref{eq: vprime of clipped relu}), %
as well as the curvature $V''_\phi(q^*)$.  Figure \ref{fig: vprimes of m} show $V'_\phi(q^*)$ and $V''_\phi(q^*)$ as functions of $m$ for different sparsity levels.  As $m$ increases, $V'_\phi(q^*)$ increases and tends to 1 from below as $m\rightarrow \infty$.  

\begin{figure}[t]
    \centering
    \begin{subfigure}[t]{0.48\textwidth}
         \centering
    \includegraphics[width=\textwidth]{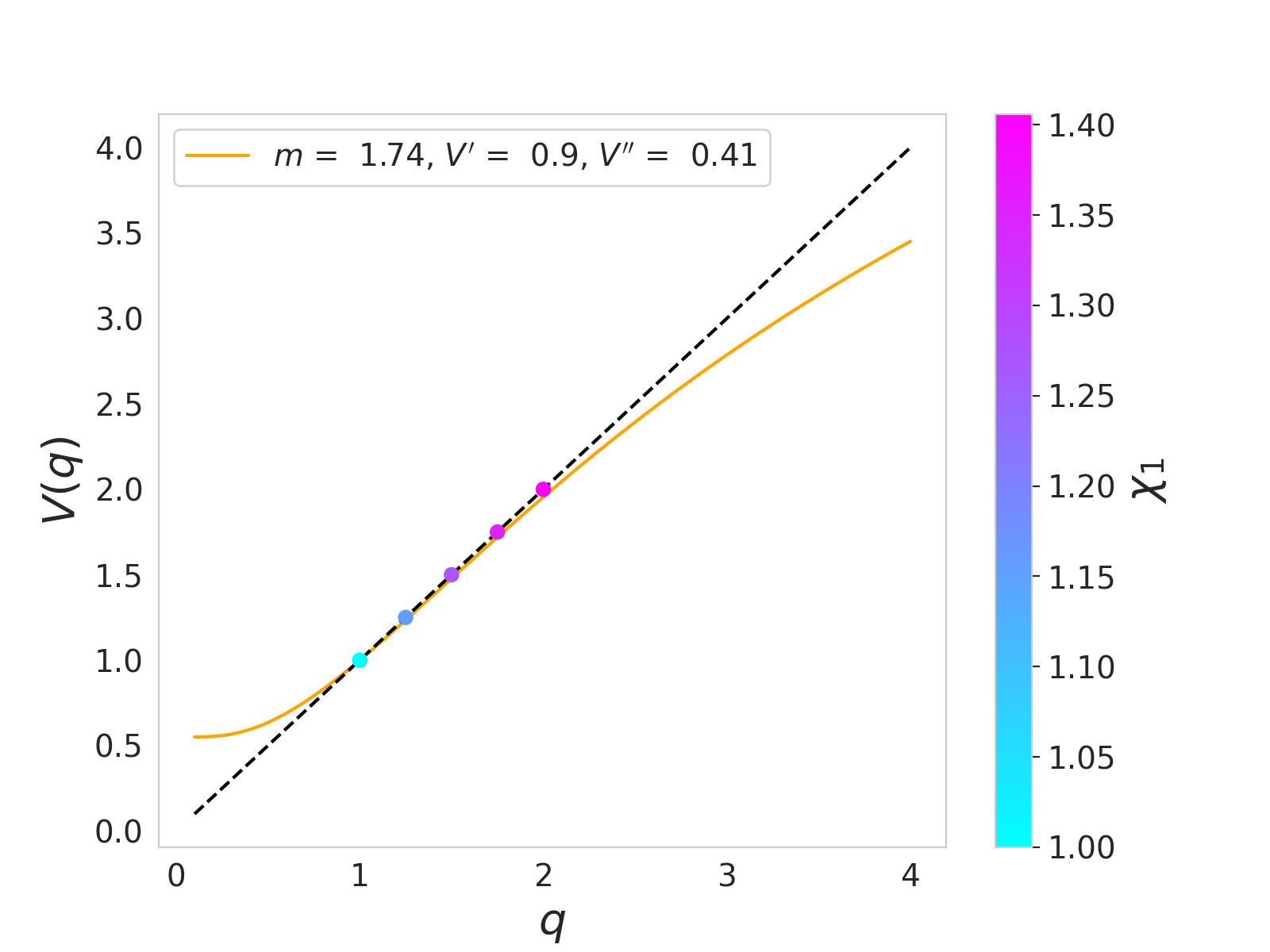}
    \caption{$\phi = \crelu$, $s=0.85$, $m$ chosen such that $V'_\phi(q^*)=0.9$ (orange), with $q^*=1$.}
     \end{subfigure}
     \hfill     
     \begin{subfigure}[t]{0.48\textwidth}
         \centering
         \includegraphics[width=\textwidth]{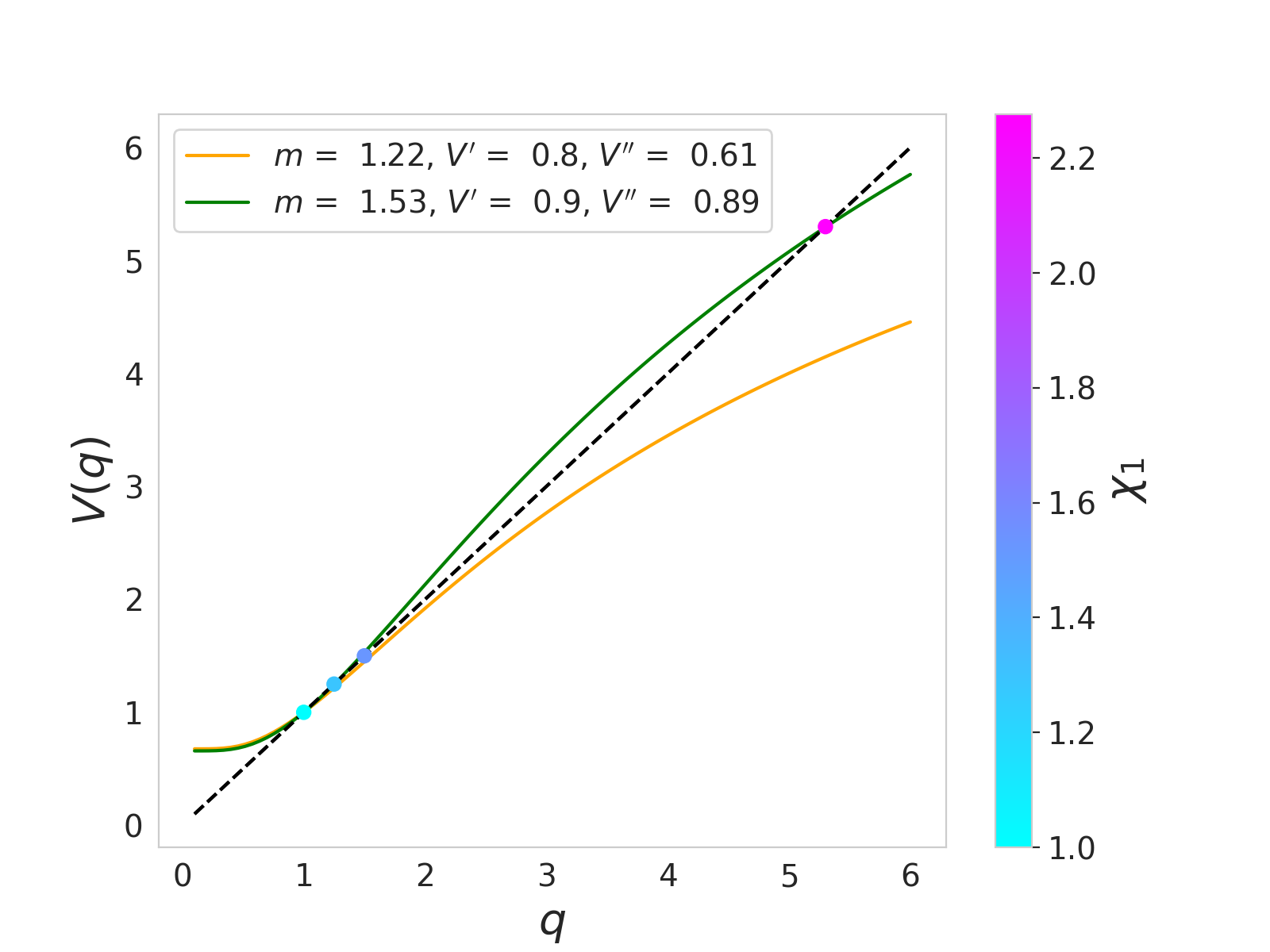}
         \caption{$\phi = \cst$, $s=0.85$, $m$ chosen such that $V'_\phi(q^*)=0.8$ (orange) and $0.9$ (green), with $q^*=1$.}
        \label{fig: st_max vmaps large vprime}
     \end{subfigure}
     \caption{Variance maps for $\crelu$ in (a) and $\cst$ in (b) at 85\% sparsity and $m$ chosen such that even though $V'_\phi(q^*)<1$ the network is unstable to train.  For $\crelu$ training fails due to $V_{\phi}(q)$ being approximately $q$ for a sufficiently large region around $q^*$ that the effective value of $\chi_1$ varies above 1 which results in  exploding gradients.  For $\cst$ training fails for the same reason of effective $\chi_1>1$, which occurs even more dramatically as $V_{\phi}(q)$ exceeds $q$ and has a second, unstable, fixed point as well as another stable fixed point, but with $\chi_1$ substantially greater than 1.}
     \label{fig: sfatrelu max vmap v'=0.9 s=0.85}
\end{figure}

While $V'_{\phi}(q^*)<1$ for each $m$, the curvature $V''_{\phi}(q^*)$ initially decreases to negative values and then increases as $m$ increases from 0. Stability for finite-dimensional networks requires $V_{\phi}(q)$ to be sufficiently stable around $q^*$ so that the natural stochasticity in $q^l$ remains in the local stable region of $V_{\phi}(q)$, meaning that $q^l$ does in practice remain approximately equal to $q^*$.  In particular, for larger values of $m$ the curvature $V''_{\phi}(q^*)$ can cause $V_{\phi}(q)$ curve up after crossing the $V_{\phi}(q)=q$ line and approximately follow it for an extended interval before flattening out again---resulting in an extended region in which $|V_{\phi}(q)-q|$ is small. Increasing $m$ (and thus $V''{\phi}(q^*)$) even further causes a bifurcation, with two new fixed points emerging: a locally unstable one close to the original $q^*$, and another locally stable one at a larger value of $q$. These new fixed points cause a failure of $q^l$ to converge to $q^*$ in practice -- instead, $q^l$ converges to or hovers at a value greater than $q^*$, corresponding to $\chi_{1,\phi} > 1$, which results in exploding gradients. See Figure \ref{fig: st_max vmaps large vprime} for illustrations of these phenomena which correspond to examples in Sec.\ \ref{sec: max variant experiments} where networks fail to train. For this reason, the values of $m$ that have the best values of accuracy and ability to train are as large as possible while $V'_{\phi}(q^*)$ and $V''_{\phi}(q^*)$ remain sufficiently small to ensure practical stability of $q^*$ based on stochasticity of $q^l$ in finite-width networks; see Table \ref{tab: results sfatrelu_max and st_max} for examples of parameters and associated test accuracy for $\crelu$ and $\st$ respectively.

\section{Experiments}\label{sec: max variant experiments}

\begin{table}[t]
    \centering
\scriptsize
\begin{tabular}{llllllcccccc}
\toprule
                         &&          &      &     &     & \multicolumn{4}{c}{DNN on MNIST} & \multicolumn{2}{c}{CNN on CIFAR10} \\
                         &&          &      &     &     & \multicolumn{2}{c}{Test accuracy} & \multicolumn{2}{c}{Test sparsity} & Test accuracy & Test sparsity \\
                         &&          &      &     &     &          mean &   std &                  mean &   std  &               &        \\
& $s$ & $\tau$ & $m$ & $V'_\phi(q^*)$ & $V''_\phi(q^*)$  &            &       &      &    &   &        \\
\midrule
$\relutau$     & 0.50 & 0.00 & N/A & 1.0 & 0.0 &       0.94 &  0.002 &                  0.50 &  0.001     &  0.70  &   0.52    \\
               & 0.60 & 0.25 & N/A & 1.0 & 0.12 &          0.76 &  0.37 &                  0.49 &  0.27     &  0.68 &   0.6    \\
               & 0.70 & 0.52 & N/A & 1.0 & 0.3 &           0.10 &  0.00 &                  0.00 &  0.00     &  0.1  &    0.0   \\
\midrule
$\st$          & 0.5 & 0.67 & N/A & 1.0 & 0.43 &           0.10 &  0.00 &                  0.00 &  0.00     &  0.1 &    0.0   \\
               & 0.6 & 0.84 & N/A & 1.0 & 0.59 &           0.10 &  0.00 &                  0.00 &  0.00     &  0.1  &   0.0    \\
               & 0.7 & 1.04 & N/A & 1.0 & 0.81 &           0.10 &  0.00 &                  0.00 &  0.00     &  0.1  &   0.0    \\
\midrule
 $\crelu$           
                    & 0.60 & 0.25 & 1.22 & 0.5 & -0.44 &          0.92 &  0.004 &                  0.60 &  0.001    & 0.70              &  0.61     \\
                        & &      & 1.63 & 0.7 & -0.42 &          0.92 &  0.003 &                  0.60 &  0.004    & 0.69             &  0.61     \\
                         &&      & 2.25 & 0.9 & -0.19 &          0.92 &  0.01 &                  0.60 &  0.01    & 0.69              &   0.60    \\
                   \cmidrule{2-12}
                    &0.70 & 0.52 & 1.05 & 0.5 & -0.37 &          0.93 &  0.01 &                  0.70 &  0.002    & 0.70              &   0.70    \\
                         &&      & 1.45 & 0.7 & -0.31 &          0.92 &  0.003 &                  0.70 &  0.002    & 0.69              &   0.69    \\
                         &&      & 2.05 & 0.9 & -0.04 &          0.92 &  0.01 &                  0.70 &  0.01    & 0.68              &   0.69    \\
                   \cmidrule{2-12}
                    & 0.80 & 0.84 & 0.89 & 0.5 & -0.24 &          0.94 &  0.004 &                  0.80 &  0.002    & 0.64              &   0.80    \\
                         & &      & 1.27 & 0.7 & -0.12 &          0.93 &  0.01 &                  0.80 &  0.01    & 0.64              &  0.78     \\
                         & &      & 1.85 & 0.9 &  0.21 &          0.94 &  0.01 &                  0.79 &  0.02    & 0.65              &   0.78    \\
                   \cmidrule{2-12}
                    & 0.85 & 1.04 & 0.81 & 0.5 & -0.14 &          0.78 &  0.16 &                  0.85 &  0.004   & 0.65              &    0.85   \\
                         & &      & 1.17 & 0.7 &  0.02 &          0.94 &  0.004 &                  0.85 &  0.003   & 0.65              &    0.84   \\
                         & &      & 1.74 & 0.9 &  0.41 &          0.28 &  0.37 &                  0.79 &  0.04   & 0.66              &    0.83   \\
\midrule
$\cst$             & 
                    0.50 & 0.67 & 0.97 & 0.5 & -0.32 &          0.92 &  0.004 &                  0.51 &  0.01      &    0.69      &   0.49    \\
                    &     &      & 1.36 & 0.7 & -0.23 &          0.92 &  0.01 &                  0.50 &  0.004      &      0.69    &   0.48    \\
                    &     &      & 1.96 & 0.9 &  0.08 &          0.92 &  0.004 &                  0.49 &  0.01      &     0.71     &   0.49    \\
                   \cmidrule{2-12}
                    & 0.60 & 0.84 & 0.89 & 0.5 & -0.24 &          0.93 &  0.01 &                  0.60 &  0.003      &    0.68      &   0.59    \\
                    &     &      & 1.27 & 0.7 & -0.12 &          0.92 &  0.004 &                  0.60 &  0.003      &     0.67     &   0.59    \\
                    &     &      & 1.85 & 0.9 &  0.21 &          0.93 &  0.004 &                  0.59 &  0.01      &    0.67      &    0.57   \\
                   \cmidrule{2-12}
                    & 0.70 & 1.04 & 0.81 & 0.5 & -0.14 &          0.94 &  0.005 &                  0.70 &  0.004      &    0.68      &   0.69    \\
                   &      &      & 1.17 & 0.7 &  0.02 &          0.93 &  0.005 &                  0.71 &  0.01      &    0.68      &   0.67    \\
                   &      &      & 1.74 & 0.9 &  0.41 &          0.27 &  0.36 &                  0.41 &  0.18      &    0.68      &   0.68    \\
                   \cmidrule{2-12}
                    & 0.80 & 1.28 & 0.72 & 0.5 &  0.00 &          0.92 &  0.04 &                  0.80 &  0.001      &     0.66     &   0.79    \\
                   &      &      & 1.06 & 0.7 &  0.23 &          0.95 &  0.01 &                  0.80 &  0.01      &     0.65     &   0.78    \\
                   &      &      & 1.61 & 0.9 &  0.69 &          0.11 &  0.01 &                  0.15 &  0.01      &      0.31    &   0.18    \\
                   \cmidrule{2-12}
                    &0.85 & 1.44 & 0.67 & 0.5 &  0.11 &          0.76 &  0.16 &                  0.85 &  0.005      &     0.63     &    0.84    \\
                   &      &      & 1.00 & 0.7 &  0.39 &          0.93 &  0.01 &                  0.85 &  0.01      &      0.63    &    0.83   \\
                   &      &      & 1.53 & 0.9 &  0.89 &          0.14 &  0.04 &                  0.11 &  0.01      &     0.29     &   0.12    \\
\bottomrule
\end{tabular}
    \caption{Experimental results for $\relutau$ and $\st$ with sparsity $s$ up to $0.7$, and for $\crelu$ and $\cst$ with sparsity $s$ up to $0.85$ and $m$ selected to have $V'_{\phi}(q^*)=0.5, 0.7,$ and $0.9$.  DNNs have 100 layers and are tested on MNIST with 5 runs and the mean and standard deviation are reported for both test accuracy and average activation sparsity calculated on the test set. For the 50-layer CNN experiments on CIFAR10 a single experiment was conducted for each hyperparameter combination.}
    \label{tab: results sfatrelu_max and st_max}
\end{table}

To confirm the theoretical predictions, we train both feedforward networks (abridged as DNNs) of width 300 and depth 100 using $\crelu$ and $\cst$ to classify digits from the MNIST dataset and, similarly, CNNs with 300 channels in each layer and depth 50 are trained to classify images from the CIFAR10 dataset. Training such deep DNNs and CNNs allows us to verify the theory, which aims to improve information propagation at initialization through many layers.  The absolute accuracy of the networks is not the focus of these experiments, rather it is the ability to retain trainability and approximately the accuracy of standard ReLU networks for sparsities greater than 50\%.  The networks are initialized at the EoC using $q^*=1$, before being trained by stochastic gradient descent (SGD) for 200 epochs with learning rate of $10^{-4}$ and $10^{-3}$ for the DNN and CNN respectively. Each experiment is conducted with sparsity $s=0.6, 0.7, 0.8,$ and $0.85$ and $m$ selected to have $V'_{\phi}(q^*)=0.5, 0.7,$ and $0.9$. The results are shown in Table~\ref{tab: results sfatrelu_max and st_max}. These results should be contrasted with those for $\relutau$ and $\st$, also shown in Table ~\ref{tab: results sfatrelu_max and st_max}, and described in App.\ \ref{sec: non max variants experiments}. 

Whereas EoC-initialized $\relutau$ and $\st$ DNNs failed to train consistently even when initialized at sparsity levels 60\% and 50\% respectively, the EoC-initialized $\crelu$ and $\cst$ DNNs can maintain full accuracy when initialized with up to 85\% sparsity. $\crelu$ and $\cst$ CNNs retain full accuracy for sparsity as high as 70\% while 85\% sparsity incurs a modest loss of accuracy to 66\%. Conversely, their non-clipped counterparts fail to train at all at 60\% and 70\% sparsity respectively. Moreover, in those cases where training is stable and high accuracy is achieved, the starting sparsity level is maintained throughout training, resulting in trained models which exhibit approximately the same high activation sparsity levels when measured on the test set. 

\begin{figure}[h]
    \centering
     \begin{subfigure}[b]{0.48\textwidth}
         \centering
         \includegraphics[width=\textwidth]{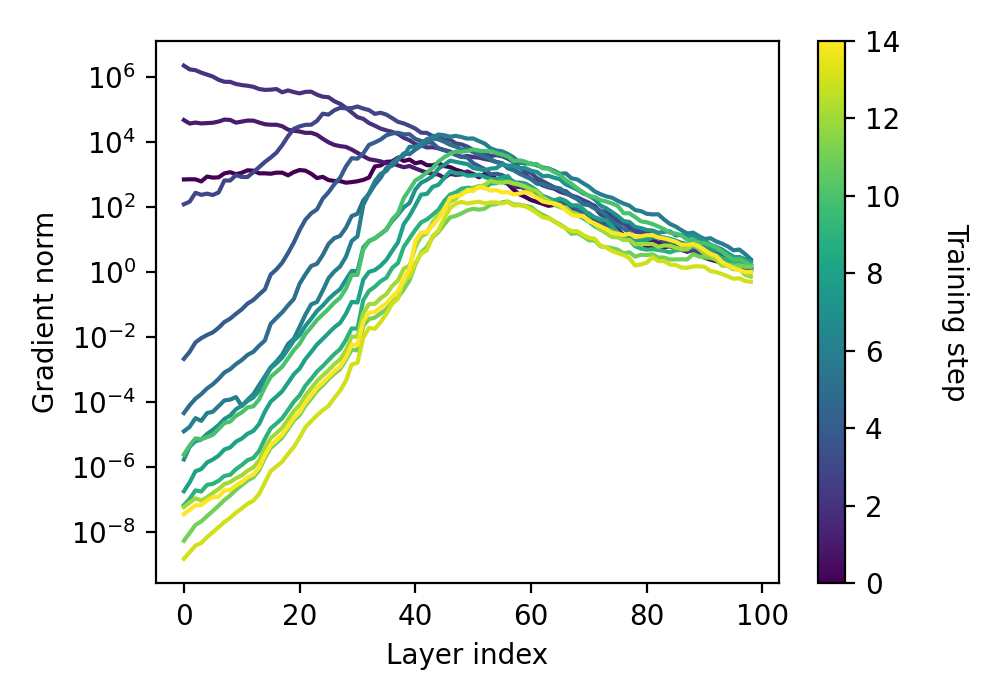}
         \caption{$\crelu$: $s=0.85$, $m=1.74$, $V'_\phi(q^*)=0.9$}
        \label{fig: sfatrelu exploding gradients}
     \end{subfigure}
     \hfill
    \begin{subfigure}[b]{0.48\textwidth}
         \centering
    \includegraphics[width=\textwidth]{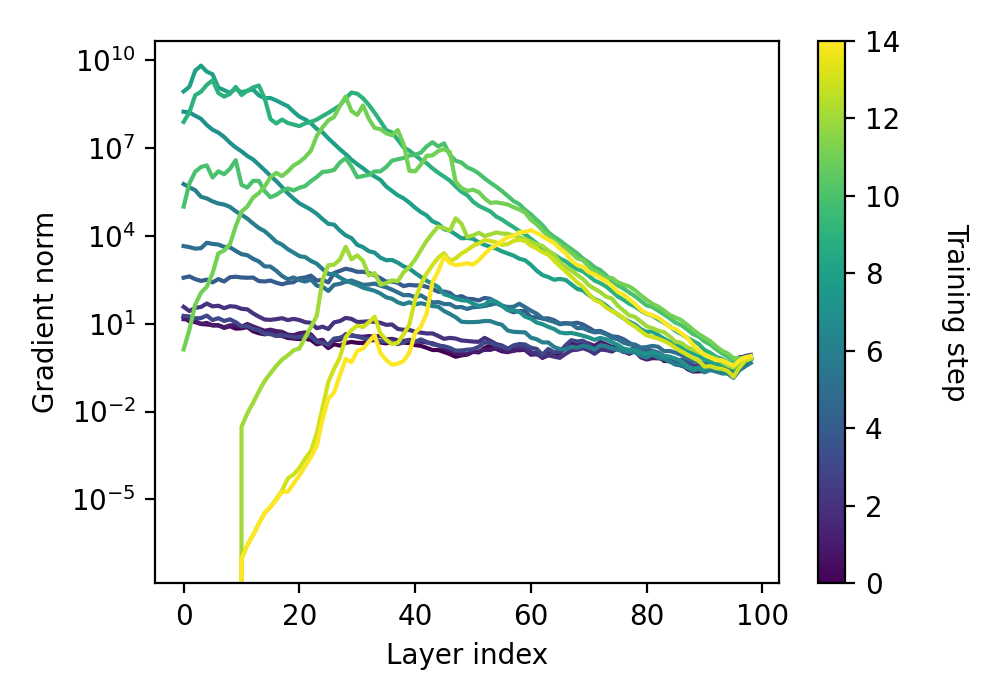}
    \caption{$\cst$: $s=0.85$, $m=1.22$, $V'_\phi(q^*)=0.8$}
    \label{fig: st exploding gradeints}
     \end{subfigure}
     \caption{Gradient norms per layer for the first 15 steps of a failed training run of DNNs on MNIST.}
     \label{fig: exploding gradients}
\end{figure}

Upon analysing the `failure' cases for DNN in which low accuracy is achieved, two failure modes are identified. The first of these corresponds to the analysis and predictions made in Section \ref{sec: choosing m}. For $\crelu$ and $s = 0.6, 0.7$, and $0.8$, we see negative or small positive values of $V''_\phi(q^*)$ for all values of $V'_\phi(q^*)$; this corresponds to a variance map at initialization with sufficient symmetry around $q^*$ to ensure that $q^l$ stays, in practice, very close to $q^*$ on average, thus avoiding any exploding gradients, and preserving trainability. Once $s$ increases to $0.85$, however, we start to observe the potential for $m$ to be too large---when $m$ is set such that $V'_\phi(q^*) = 0.9$, then $V''_\phi(q^*) = 0.41$, a high chance of failure in training is observed for the reasons described in Section \ref{sec: choosing m} and highlighted by the variance map in Figure \ref{fig: sfatrelu max vmap v'=0.9 s=0.85}. Indeed, training fails in four out of five experiments, reflected by the low mean test accuracy and high standard deviation. Figure \ref{fig: sfatrelu exploding gradients} plots the gradient norms at each layer for the first 15 steps of training for one of these failed runs. Even at the very first step, the gradients grow as the backwards pass proceeds (up to $O(10^3)$), and this reaches $O(10^6)$ by step 3\footnote{The reason for the drop back to very low levels at later steps is due to the fact that once the weights (and thus activations) grow large enough, they almost all begin to land in the zero-gradient interval of the activation functions, resulting in no further gradient flow through those layers.}.  

The same trend is observed for DNNs and $s\geq 0.8$ with $\phi = \cst$. Starting from the lowest tested value of $m$ in these cases,  increasing $m$ initially increases test accuracy, but at a certain point, $m$ becomes too large, such that the corresponding combination of $V'_\phi(q^*)$ and $V''_\phi(q^*)$ cause failure in some (and then all) of the five experiments.  Figure \ref{fig: st exploding gradeints} shows the layerwise gradient norms over the first 15 training steps of a failed run with $\cst$ with $s=0.85$ and $m=1.22$. Though on step 1 the gradients remain $\sim O(1)$, very quickly the exploding gradient phenomenon emerges.

In contrast, Table \ref{tab: results sfatrelu_max and st_max} shows that CNNs with $\crelu$ exhibit a more stable test accuracy, whereas the analogous tests for $\cst$ show a loss of accuracy for large $m$ where $V'_{\phi}(q^*)=0.9$ once $V''_{\phi}(q^*)\ge 0.69$ at $s=0.8$ and $m=1.61$ as well as $s=0.85$ and $m=1.53$.

The second failure mode accounts for the failure to train to high accuracy once starting sparsity reaches 85\%  for low values of $m$ (the failure when $s=0.85$ and $m$ is large is captured in the first failure mode); specifically, the failure of DNNs to obtain high accuracy when $s=0.85$ and $m=0.81$ for $\crelu$, as well as when $s=0.85$ and $m\leq0.83$ for $\cst$. In all of these cases, EoC initialization avoids exploding gradients at initialization and performance does improve during training for all of the five runs. However, the resulting accuracy is reduced, reaching a maximum of $\approx$78\% when $s=0.85$ at $m=0.81$, for example. This is discussed further in App.\ \ref{sec: loss function}.

One final observation from Table \ref{tab: results sfatrelu_max and st_max} is that for DNNs the maximum accuracy actually increases with increasing sparsity (up to a point, and for sufficiently large values of $m$). Maximum accuracy for $\relutau$ networks is achieved when sparsity is 80\% and 85\% (accuracy of 94\%), and for $\cst$ networks when sparsity is 80\% (accuracy of 95\%). %
However, CNNs in contrast do not show the aforementioned improved accuracy as a result of sparsity, instead they exhibit a more gradual loss of accuracy and train for a somewhat larger range of $m$.

\section{Conclusions and avenues for future work}\label{sec:conclusions}
The analysis here explains why arguably the most natural highly-sparsifying activation functions $\relutau$ and $\st$ fail to train for deep DNNs and CNNs.  This instability is then overcome with a simple modification in the form of a bound on the absolute value of these functions, yielding a stable EoC initializations for the resulting networks, which allows for training very deep networks with these modified activation functions with activation sparsity up to 85\%. 

There are multiple natural and promising avenues to extend this line of work. Firstly, a more complete explanation of the unstable training dynamics when $s$ is very large and $m$ is small (the second failure mode identified in Section \ref{sec: max variant experiments}). Secondly, extension of EoC analysis for these sparsifying activation functions from DNNs and CNNs to transformer architectures, and networks with skip connections like ResNet would be important to have the greatest impact in recent applications.  Additionally, it would be interesting to investigate whether smooth variants of $\crelu$ and or $\cst$ exhibit any preferable properties over their piecewise-linear variants, specifically through controlling higher order moments of the spectrum of the input-output Jacobian.  Finally, derivations of formulae for the largest choice of $m$ for which the variance map is stable to the variance of $q^l$ determined by the network width, input channels or other numbers of network parameters.

\section*{Reproducibility}
Derivations of the theory in Section \ref{sec:instability} and \ref{sec:clipped} can be found in Appendices \ref{sec: activation plots} and \ref{sec: clipped plots}. Experimental setup and hyperparamters are detailed in Section \ref{sec: max variant experiments} and Appendices \ref{sec: non max variants experiments} and \ref{sec:additional experimental details}.

\bibliography{main.bib}

\begin{thebibliography}{25}
\providecommand{\natexlab}[1]{#1}
\providecommand{\url}[1]{\texttt{#1}}
\expandafter\ifx\csname urlstyle\endcsname\relax
  \providecommand{\doi}[1]{doi: #1}\else
  \providecommand{\doi}{doi: \begingroup \urlstyle{rm}\Url}\fi

\bibitem[Bizopoulos \& Koutsouris(2021)Bizopoulos and Koutsouris]{sparse_activated}
Paschalis Bizopoulos and Dimitrios Koutsouris.
\newblock Sparsely activated networks.
\newblock \emph{IEEE Transactions on Neural Networks and Learning Systems}, 32\penalty0 (3):\penalty0 1304--1313, 2021.
\newblock \doi{10.1109/TNNLS.2020.2984514}.

\bibitem[Blalock et~al.(2020)Blalock, Gonzalez~Ortiz, Frankle, and Guttag]{Blalock2020}
Davis Blalock, Jose~Javier Gonzalez~Ortiz, Jonathan Frankle, and John Guttag.
\newblock What is the state of neural network pruning?
\newblock In I.~Dhillon, D.~Papailiopoulos, and V.~Sze (eds.), \emph{Proceedings of Machine Learning and Systems}, volume~2, pp.\  129--146, 2020.
\newblock URL \url{https://proceedings.mlsys.org/paper_files/paper/2020/file/6c44dc73014d66ba49b28d483a8f8b0d-Paper.pdf}.

\bibitem[Chen et~al.(2018)Chen, Pennington, and Schoenholz]{Chen_2018}
Minmin Chen, Jeffrey Pennington, and Samuel Schoenholz.
\newblock Dynamical isometry and a mean field theory of {RNN}s: Gating enables signal propagation in recurrent neural networks.
\newblock In Jennifer Dy and Andreas Krause (eds.), \emph{Proceedings of the 35th International Conference on Machine Learning}, volume~80 of \emph{Proceedings of Machine Learning Research}, pp.\  873--882. PMLR, 10--15 Jul 2018.
\newblock URL \url{https://proceedings.mlr.press/v80/chen18i.html}.

\bibitem[Donoho(1995)]{denoise}
D.L. Donoho.
\newblock De-noising by soft-thresholding.
\newblock \emph{IEEE Transactions on Information Theory}, 41\penalty0 (3):\penalty0 613--627, 1995.
\newblock \doi{10.1109/18.382009}.

\bibitem[Foucart \& Rauhut(2013)Foucart and Rauhut]{foucart_rauhut}
Simon Foucart and Holger Rauhut.
\newblock \emph{A Mathematical Introduction to Compressive Sensing.}
\newblock Applied and Numerical Harmonic Analysis. Birkhäuser, 2013.
\newblock ISBN 978-0-8176-4947-0.

\bibitem[Georgiadis(2019)]{georgiadis2019accelerating}
Georgios Georgiadis.
\newblock Accelerating convolutional neural networks via activation map compression.
\newblock In \emph{2019 IEEE/CVF Conference on Computer Vision and Pattern Recognition (CVPR)}, pp.\  7078--7088, 2019.
\newblock \doi{10.1109/CVPR.2019.00725}.

\bibitem[Gilboa et~al.(2019)Gilboa, Chang, Chen, Yang, Schoenholz, Chi, and Pennington]{Gilboa_2019}
Dar Gilboa, Bo~Chang, Minmin Chen, Greg Yang, Samuel~S. Schoenholz, Ed~H. Chi, and Jeffrey Pennington.
\newblock Dynamical isometry and a mean field theory of lstms and grus, 2019.
\newblock URL \url{https://arxiv.org/abs/1901.08987}.

\bibitem[Han et~al.(2015)Han, Pool, Tran, and Dally]{han2015learning}
Song Han, Jeff Pool, John Tran, and William Dally.
\newblock Learning both weights and connections for efficient neural network.
\newblock In C.~Cortes, N.~Lawrence, D.~Lee, M.~Sugiyama, and R.~Garnett (eds.), \emph{Advances in Neural Information Processing Systems}, volume~28. Curran Associates, Inc., 2015.
\newblock URL \url{https://proceedings.neurips.cc/paper_files/paper/2015/file/ae0eb3eed39d2bcef4622b2499a05fe6-Paper.pdf}.

\bibitem[Hayou et~al.(2020)Hayou, Ton, Doucet, and Teh]{hayou2020robust}
Soufiane Hayou, Jean-Francois Ton, Arnaud Doucet, and Yee~Whye Teh.
\newblock Robust pruning at initialization.
\newblock In \emph{International Conference on Learning Representations}, 2020.

\bibitem[Hayou et~al.(2021)Hayou, Ton, Doucet, and Teh]{Hayou_2020}
Soufiane Hayou, Jean-Francois Ton, Arnaud Doucet, and Yee~Whye Teh.
\newblock Robust pruning at initialization.
\newblock In \emph{International Conference on Learning Representations}, 2021.
\newblock URL \url{https://openreview.net/forum?id=vXj_ucZQ4hA}.

\bibitem[Hoefler et~al.(2021)Hoefler, Alistarh, Ben-Nun, Dryden, and Peste]{hoefler2021sparsity}
Torsten Hoefler, Dan Alistarh, Tal Ben-Nun, Nikoli Dryden, and Alexandra Peste.
\newblock Sparsity in deep learning: Pruning and growth for efficient inference and training in neural networks.
\newblock \emph{The Journal of Machine Learning Research}, 22\penalty0 (1):\penalty0 10882--11005, 2021.

\bibitem[Huang et~al.(2020)Huang, Xu, Du, Zeng, and Zhao]{Huang_2019}
Wei Huang, Richard Yi~Da Xu, Weitao Du, Yutian Zeng, and Yunce Zhao.
\newblock Mean field theory for deep dropout networks: Digging up gradient backpropagation deeply.
\newblock In Giuseppe~De Giacomo, Alejandro Catal{\'{a}}, Bistra Dilkina, Michela Milano, Sen{\'{e}}n Barro, Alberto Bugar{\'{\i}}n, and J{\'{e}}r{\^{o}}me Lang (eds.), \emph{{ECAI} 2020 - 24th European Conference on Artificial Intelligence, 29 August-8 September 2020, Santiago de Compostela, Spain, August 29 - September 8, 2020 - Including 10th Conference on Prestigious Applications of Artificial Intelligence {(PAIS} 2020)}, volume 325 of \emph{Frontiers in Artificial Intelligence and Applications}, pp.\  1215--1222. {IOS} Press, 2020.
\newblock \doi{10.3233/FAIA200221}.
\newblock URL \url{https://doi.org/10.3233/FAIA200221}.

\bibitem[Kurtz et~al.(2020)Kurtz, Kopinsky, Gelashvili, Matveev, Carr, Goin, Leiserson, Moore, Shavit, and Alistarh]{kurtz2020inducing}
Mark Kurtz, Justin Kopinsky, Rati Gelashvili, Alexander Matveev, John Carr, Michael Goin, William Leiserson, Sage Moore, Nir Shavit, and Dan Alistarh.
\newblock Inducing and exploiting activation sparsity for fast inference on deep neural networks.
\newblock In \emph{International Conference on Machine Learning}, pp.\  5533--5543. PMLR, 2020.

\bibitem[Lee et~al.(2018)Lee, Sohl-dickstein, Pennington, Novak, Schoenholz, and Bahri]{Lee_2017}
Jaehoon Lee, Jascha Sohl-dickstein, Jeffrey Pennington, Roman Novak, Sam Schoenholz, and Yasaman Bahri.
\newblock Deep neural networks as gaussian processes.
\newblock In \emph{International Conference on Learning Representations}, 2018.
\newblock URL \url{https://openreview.net/forum?id=B1EA-M-0Z}.

\bibitem[Lee et~al.(2019)Lee, Ajanthan, and Torr]{lee2019snip}
Namhoon Lee, Thalaiyasingam Ajanthan, and Philip Torr.
\newblock Snip: Single-shot network pruning based on connection sensitivity.
\newblock In \emph{International Conference on Learning Representations}, 2019.
\newblock URL \url{https://openreview.net/forum?id=B1VZqjAcYX}.

\bibitem[Murray et~al.(2022)Murray, Abrol, and Tanner]{murray2022activation}
Michael Murray, Vinayak Abrol, and Jared Tanner.
\newblock Activation function design for deep networks: linearity and effective initialisation.
\newblock \emph{Applied and Computational Harmonic Analysis}, 59:\penalty0 117--154, 2022.

\bibitem[Pennington et~al.(2017)Pennington, Schoenholz, and Ganguli]{pennington2017resurrecting}
Jeffrey Pennington, Samuel Schoenholz, and Surya Ganguli.
\newblock Resurrecting the sigmoid in deep learning through dynamical isometry: theory and practice.
\newblock \emph{Advances in neural information processing systems}, 30, 2017.

\bibitem[Pennington et~al.(2018)Pennington, Schoenholz, and Ganguli]{pennington2018emergence}
Jeffrey Pennington, Samuel Schoenholz, and Surya Ganguli.
\newblock The emergence of spectral universality in deep networks.
\newblock In \emph{International Conference on Artificial Intelligence and Statistics}, pp.\  1924--1932. PMLR, 2018.

\bibitem[Poole et~al.(2016)Poole, Lahiri, Raghu, Sohl-Dickstein, and Ganguli]{poole2016exponential}
Ben Poole, Subhaneil Lahiri, Maithra Raghu, Jascha Sohl-Dickstein, and Surya Ganguli.
\newblock Exponential expressivity in deep neural networks through transient chaos.
\newblock \emph{Advances in neural information processing systems}, 29, 2016.

\bibitem[Roberts et~al.(2022)Roberts, Yaida, and Hanin]{PDLT-2022}
Daniel~A. Roberts, Sho Yaida, and Boris Hanin.
\newblock \emph{The Principles of Deep Learning Theory}.
\newblock Cambridge University Press, 2022.
\newblock \url{https://deeplearningtheory.com}.

\bibitem[Schoenholz et~al.(2017)Schoenholz, Gilmer, Ganguli, and Sohl-Dickstein]{schoenholz2017deep}
Samuel~S. Schoenholz, Justin Gilmer, Surya Ganguli, and Jascha Sohl-Dickstein.
\newblock Deep information propagation.
\newblock In \emph{International Conference on Learning Representations}, 2017.
\newblock URL \url{https://openreview.net/forum?id=H1W1UN9gg}.

\bibitem[Tanaka et~al.(2020)Tanaka, Kunin, Yamins, and Ganguli]{tanaka2020pruning}
Hidenori Tanaka, Daniel Kunin, Daniel~L Yamins, and Surya Ganguli.
\newblock Pruning neural networks without any data by iteratively conserving synaptic flow.
\newblock \emph{Advances in neural information processing systems}, 33:\penalty0 6377--6389, 2020.

\bibitem[Xiao et~al.(2018)Xiao, Bahri, Sohl-Dickstein, Schoenholz, and Pennington]{Xiao_2018}
Lechao Xiao, Yasaman Bahri, Jascha Sohl-Dickstein, Samuel Schoenholz, and Jeffrey Pennington.
\newblock Dynamical isometry and a mean field theory of {CNN}s: How to train 10,000-layer vanilla convolutional neural networks.
\newblock In Jennifer Dy and Andreas Krause (eds.), \emph{Proceedings of the 35th International Conference on Machine Learning}, volume~80 of \emph{Proceedings of Machine Learning Research}, pp.\  5393--5402. PMLR, 10--15 Jul 2018.
\newblock URL \url{https://proceedings.mlr.press/v80/xiao18a.html}.

\bibitem[Yang \& Schoenholz(2017)Yang and Schoenholz]{Yang_2017}
Ge~Yang and Samuel Schoenholz.
\newblock Mean field residual networks: On the edge of chaos.
\newblock In I.~Guyon, U.~Von Luxburg, S.~Bengio, H.~Wallach, R.~Fergus, S.~Vishwanathan, and R.~Garnett (eds.), \emph{Advances in Neural Information Processing Systems}, volume~30. Curran Associates, Inc., 2017.
\newblock URL \url{https://proceedings.neurips.cc/paper_files/paper/2017/file/81c650caac28cdefce4de5ddc18befa0-Paper.pdf}.

\bibitem[Yang et~al.(2019)Yang, Pennington, Rao, Sohl-Dickstein, and Schoenholz]{Yang_2019}
Greg Yang, Jeffrey Pennington, Vinay Rao, Jascha Sohl-Dickstein, and Samuel~S. Schoenholz.
\newblock A mean field theory of batch normalization.
\newblock In \emph{International Conference on Learning Representations}, 2019.
\newblock URL \url{https://openreview.net/forum?id=SyMDXnCcF7}.

\end{thebibliography}
\bibliographystyle{iclr2024_conference}

\newpage
\appendix
\section{Gaussian process: variance, covariance, and gradient propagation at initialization}\label{sec: background}

Numerous prior works \cite{poole2016exponential, schoenholz2017deep, pennington2017resurrecting, murray2022activation} have considered how the hyperparameters $\sigma_w^2$ and $\sigma_b^2$, and the choice of activation function $\phi$, affect signal and information propagation through the network at initialization, as well as the network's trainability via gradient backpropagation. The standard approach in these works is to perform a mean-field analysis in which one considers each layer to have infinite width, that is, $N^{l} \longrightarrow \infty$ for all $l$. In particular \citep{poole2016exponential} considered the (width-scaled) length of the pre-activations.

The shape of both $V_\phi$ and $R_\phi$, along with the number of fixed points they admit and their associated stabilities depend on $\phi$, $\sigma_w^2$ and $\sigma_b^2$.  Of particular interest is the slope of this correlation map at $\rho^* = 1$, which is given by 
\begin{align}
    \chi_{1,\phi} := R_\phi'(1) = \sigma_w^2 \int_\R \left(\phi'(\sqrt{q^*}\, z )\right)^2 \gamma(dz).
\end{align}
When $\chi_{1,\phi}<1$, the fixed point $\rho^*$ is stable, and any pair of points will asymptotically converge exponentially fast to being exactly aligned, whereas when $\chi_{1,\phi}>1$, the fixed point $\rho^*$ is unstable, and another fixed point emerges $\rho^* < 1$. In the latter case, even two input points which are arbitrarily close to one another will asymptotically decorrelate. The former $\chi_{1,\phi}<1$ regime is referred to in the literature as the `ordered' regime, and the latter $\chi_{1,\phi}>1$ regime is dubbed the `chaos' regime. The `Edge of Chaos' (EoC) is the set of points $(\sigma_w,\sigma_b)$ which separate these two regions, that is,  the set of points $(\sigma_w,\sigma_b)$ such that $V(q^*) = q^*$ and $\chi_{1,\phi} = 1$.

It has been proven that initialising a network at the Edge of Chaos improves the depth of information propagation at initialization \cite{Yang_2017}. Moreover, as shown in \cite{schoenholz2017deep}, EoC initialization also helps to prevent exploding and vanishing gradients early in training, which otherwise make a network untrainable. To understand why this is the case, consider the backpropagation algorithm with which deep networks are trained via (stochastic) gradient-based optimisation methods. Given a loss function $\mathcal{L}(x, \theta)$ where $\theta = \bigcup_{l=1}^L \{W^l\} \cup\{b^l\}$, backpropagation calculates the gradients with respect to the weights and biases according to the follow recurrence relations
\begin{align}
    \frac{\partial  \mathcal{L}(x, \theta)}{\partial h^L} & := \delta^L = D^L\nabla_{h^l}\mathcal{L}(x, \theta), \\
     \frac{\partial  \mathcal{L}(x, \theta)}{\partial h^l} & := \delta^l = (D^lW^{l+1})^\top \delta^{l+1}, \label{eq: backprop linear operator} \\
    \frac{\partial  \mathcal{L}(x, \theta)}{\partial b_i^l} & = \delta_i^l, \\
    \frac{\partial  \mathcal{L}(x,\theta)}{\partial w_{i, j}^l} & = h_j^{l-1} \delta_i^l,
\end{align}
where $D^l$ is the diagonal matrix with entries $D^l_{ii} = \phi'(h_i^l)$. In \cite{pennington2018emergence} it was shown that in the infinite-width limit, though error vectors $\delta^l$ are not Gaussian-like $h^l$, the second moment of the error vectors distribution evolves according to its own recurrence relation which depends on the value of $\chi_{1,\phi}$. Specifically, they show that (given the simplifying assumption that weights used during forward and backward passes are drawn independently from each other) $\tilde{q}^l:= \mathbb{E}[(\delta_i^l)^2]$ evolves according to
\begin{align}\label{eq: error variance relation}
    \tilde{q}^l = \tilde{q}^{l+1}\frac{N_{l+1}}{N_l} \chi_{1,\phi}.
\end{align}
Thus when $\chi_{1,\phi}>1$, the gradient norm is expected to increase exponentially over the course of the backward pass (a phenomenon known as `exploding gradients'), and conversely to shrink exponentially (known as `vanishing gradients') when $\chi_{1,\phi}<1$. Both of these phenomena impede successful training of the network. In contrast, when $\chi_{1,\phi}=1$ (and all layers have the equal width), gradient norm is preserved in expectation from one layer to the next, which allows for meaningful updates to be applied to all parameters. 

The $\chi_{1,\phi}=1$ condition also emerges from the analysis of the input-output Jacobian using tools from free probability in \cite{pennington2017resurrecting}. The input-output Jacobian $J$ is given by
\begin{align}
    J = \frac{\partial x^L}{\partial x^0} = \prod_{l=1}^L D^lW^l.
\end{align}
The Jacobian is (as expected) closely related to the operations involved in backpropagation - and more precisely involves the product of the linear backpropagation operators applied to compute each layer's error vector during the backwards pass (see Equation \eqref{eq: backprop linear operator}). In \cite{pennington2018emergence}, the authors consider the limiting spectral density of $JJ^\top$ with moment generating function
\begin{align}
    M_{JJ^\top}(z) = \sum_{k=1}^\infty \frac{m_k}{z^k}.
\end{align}
To briefly recap the key points of the analysis in \cite{pennington2018emergence}, assuming normalised input such that $q^1 = q^*$, then the distribution of $D^l$ is independent of $l$, and the moments of $D$ are given by
\begin{align}\label{eq: mu_k}
    \mu_k = \int (\phi'(\sqrt{q^*} z ))^{2k} \gamma(dz). 
\end{align}
The first two of these ($\mu_1$ and $\mu_2$)  define the first two moments of the spectrum of $JJ^\top$, $m_1$ and $m_2$, according to 
\begin{align}
    m_1 &= (\sigma_w^2 \mu_1)^L = (\chi_{1,\phi})^L \label{eq: JJ^T m1}\\
    m_2 &= (\sigma_w^2 \mu_1)^{2L} L\left( \frac{\mu_1}{\mu_2^2} + \frac{1}{L} - 1 - s_1 \right),\label{eq: JJ^T m2}
\end{align}
where $s_1$ is the first moment of the $S$-transform of $WW^\top$. When $W$ is Gaussian with mean 0 and variance $\sigma_w^2/N_{l-1}$ (as in the setup considered here), $s_1 = - 1$ . In  \cite{pennington2018emergence}, the authors also consider the case when $W$ is orthogonal, in which case $s_1=0$. 

From Equation \eqref{eq: JJ^T m1}, $m_1 = (\chi_{1,\phi})^L$, which implies that for large $L$, the scale of the error vectors in early layers is likely to blow up or collapse to 0 if $\chi_{1,\phi}>1$ or $\chi_{1,\phi}<1$ respectively. However, while this $\chi_{1,\phi}=1$ condition stops the growth of $m_1$ as $L$ grows large, this only tells us about the growth in error vector in expectation. In \cite{pennington2018emergence}, the authors' key motivation is to investigate the possibility of imposing a stronger condition: (approximate) dynamical isometry, where all singular values of the Jacobian are concentrated around one, protecting against even the worst case growth and decay in the error signal.

Equations \eqref{eq: JJ^T m1} and \eqref{eq: JJ^T m2} yield an expression for the variance of the spectrum of $JJ^\top$ when $\chi_{1,\phi} = 1$,
\begin{align}\label{eq: JJ^T var}
    \sigma_{JJ^\top}^2 = m_2 - m_1^2 = L \left( \frac{\mu_1}{\mu_2^2} - 1 - s_1  \right).
\end{align}

The variance of the spectrum of $JJ^\top$ for both $\relutau$ and $\st$ can be readily calculated using \eqref{eq: mu_k} in App.\ \ref{sec: clipped jacobian} which grows linearly with depth as $L/(1-s)$ and $Ls/(1-s)$ for Gaussian and orthogonal weights respectively.

This shows that even with $\chi_{1,\phi}=1$, the variance of the Jacobian spectrum grows linearly with depth $L$ for generic $\mu_k$ and $s_1$. Whether or not this is possible to avoid depends on the activation function and weights initialisaton scheme. One case analysed in \cite{pennington2018emergence}, for example, is the HardTanh activation for which  $\sigma_w^2(q^*) \longrightarrow 1$ as $q^* \longrightarrow 0$. Together with Equation \eqref{eq: JJ^T var}, this means that while for Gaussian weights, with $s_1 = -1$,  $\sigma_{JJ^\top}^2 \propto L$ for all $q^*$, with orthogonal weights ($s_1=0$) one has that 
\begin{align}
    \sigma_{JJ^\top}^2 \longrightarrow 0 \qquad \text{as} \qquad q^* \longrightarrow 0, 
\end{align}
for fixed $L$, meaning that one can arbitrarily shrink the variance of the spectrum of the Jacobian by sufficiently shrinking $q^*$. They show similarly that it is possible to achieve arbitrarily small $\sigma_{JJ^\top}^2$ for networks with Erf activation functions and orthogonal (but not Gaussian) weights initializations, and also that this is not possible for ReLU networks (neither with Gaussian nor orthogonal weights).  

Empirically, even in the absence of dynamical isometry, EoC initialization yields significant benefits in terms of enabling and speeding up initial training of very deep nets \cite{pennington2017resurrecting}. 

The process for calculating EoC $\sigma_w^2$ and $\sigma_b^2$ is as follows: (i) select a $q^*$, say the variance of your (typically already normalised) input data; (ii) use Equation~\eqref{eq: chi1} with $\chi_{1,\phi} = 1$ to solve for $\sigma_w^2$; (iii) use Equation~\eqref{eq: variance map} with $V(q^*) = q^*$ to solve for $\sigma_b^2$.

\section{Variance and correlation maps for $\relutau$ and $\st$} \label{sec: activation plots}

Recall that the variance map for a general activation function $\phi$ is the following:
\begin{equation}
    V_\phi(q) = \sigma^2_w \int_\R (\phi(\sqrt{q} z))^2 \, \gamma(dz) + \sigma^2_b,
\end{equation}
where $\gamma(dz) = \exp(-z^2/2)/\sqrt{2\pi} \, dz$ is the standard normal distribution. We define 
\begin{equation}
\Phi(z) := \int_{-\infty}^z \, \gamma(dt)
\end{equation} 
being the cumulative distribution function (CDF) of a standard normal distribution, which is related to the error function by 
\begin{gather}
\text{erf}(x) = 2\Phi(\sqrt{2} \, x) - 1 \iff \Phi(\sqrt{2}\,x) = \frac{\text{erf}(x)+1}{2}, \\
\text{erf}^{-1}(x) = \frac{1}{\sqrt{2}}\Phi^{-1}\bracket{\frac{x+1}{2}} \iff \Phi^{-1}(x) = \sqrt{2} \text{erf}^{-1}(2x-1).
\end{gather}

Recall further that $ \Phi(-z) = 1 - \Phi(z)$, as well as the following identities:
\begin{equation}
    \int_z^\infty t \, \gamma(dt) = \frac{\exp(-z^2/2)}{\sqrt{2\pi}}, \quad \int_z^\infty t^2 \, \gamma(dt) = z\frac{\exp(-z^2/2)}{\sqrt{2\pi}} + (1-\Phi(z)).
\end{equation}
With these in mind, the variance map for $\phi = \relutau$ can be computed as follows:
\begin{align*}
V_{\relutau}(q) &= \sigma^2_w \int^\infty_{\tau/\sqrt{q}} (\sqrt{q}z - \tau)^2 \, \gamma(dz) + \sigma^2_b \\
&= \sigma^2_w \sqbracket{\int_{\tau/\sqrt{q}}^\infty (qz^2 - 2\sqrt{q}\tau z + \tau^2) \, \gamma(dz)} + \sigma^2_b \\
&= \sigma^2_w \sqbracket{q\int_{\tau/\sqrt{q}}^\infty z^2 \, \gamma(dz) - 2\sqrt{q}\tau \int_{\tau/\sqrt{q}}^\infty z \, \gamma(dz) + \tau^2 \int_{\tau/\sqrt{q}}^\infty \gamma(dz)} + \sigma^2_b \\
&= \sigma^2_w \Bigg[q\bracket{\frac{\tau}{\sqrt{2\pi q}} \exp\bracket{-\frac{\tau^2}{2q}} + 1 - \Phi \bracket{\frac{\tau}{\sqrt{q}}}} \\
&\phantom{=}- \frac{2\sqrt{q}\tau}{\sqrt{2\pi}} \exp\bracket{-\frac{\tau^2}{2q}} + \tau^2 \bracket{1 - \Phi\bracket{\frac{\tau}{\sqrt{q}}}} \Bigg] + \sigma^2_b \\
&= \sigma^2_w \sqbracket{(q + \tau^2) \bracket{ 1-\Phi\bracket{\frac{\tau}{\sqrt{q}}}} - \frac{\sqrt{q}\tau}{\sqrt{2\pi}} \exp\bracket{-\frac{\tau^2}{2q}}} + \sigma_b^2. \numberthis
\end{align*}

Therefore,
\begin{align*}
V'_{\relutau}(q) &= \sigma^2_w \Bigg[\tilde{\Phi}\bracket{\frac{\tau}{\sqrt{q}}} - \frac{\tau(q+\tau^2)}{2q^{3/2}\sqrt{2\pi}} \exp\bracket{-
\frac{\tau^2}{2q}} - \frac{\tau}{2\sqrt{2\pi q}} \exp\bracket{-\frac{\tau^2}{2q}} \\
&\phantom{=}- \sqrt{q} \tau \bracket{\frac{\tau}{\sqrt{2 \pi q}}} \exp\bracket{-\frac{\tau^2}{2q}} \frac{\tau}{2q^{3/2}} \Bigg] \\
&= \sigma^2_w \sqbracket{1-\Phi\bracket{\frac{\tau}{\sqrt{q}}}} = \sigma^2_w \Phi\bracket{-\frac{\tau}{\sqrt{q}}}. \numberthis
\end{align*}

Let $q^*$ be a fixed point of the variance map, such that $V_{\relutau}(q^*) = q^*$. The slope of the correlation map at $\rho = 1$ is then given by
\begin{align*}
\chi_{1, \relutau} &= \sigma^2_w \int_\R \bracket{\phi'(\sqrt{q^*} z)}^2 \, \gamma(dz) \\
&= \sigma^2_w \int_{\tau/\sqrt{q^*}}^\infty \, \gamma(dz) = \sigma^2_w \sqbracket{1 -\Phi\bracket{\frac{\tau}{\sqrt{q^*}}}} = V'_{\relutau}(q^*). \numberthis
\end{align*}
As a result, if a DNN is initialized at the EoC (i.e. $\chi_{1, \relutau} = 1$), then the fixed point of the variance map satisfies $V'_{\relutau}(q^*) = \chi_{1,\relutau} = 1$. Moreover, note that for all $q$,
\begin{align}
    V''_{\relutau}(q) = \sigma^2_w \frac{\tau}{2\sqrt{2\pi} q^{3/2}} \exp\bracket{-\frac{\tau^2}{2q}} > 0.
\end{align}
This shows that the fixed point $q^*$ is unstable when the DNN is initialized at the EoC.

This phenomenon also holds for DNNs with soft thresholding activation function $\st$. In fact, 
\begin{align*}
V_{\st}(q) &= \sigma^2_w \sqbracket{\int_{-\infty}^{-\tau/\sqrt{q}} (\sqrt{q} z + \tau)^2 \, \gamma(dz) + \int^\infty_{\tau/\sqrt{q}} (\sqrt{q}z - \tau)^2 \, \gamma(dz)} + \sigma^2_b \\
&= 2\sigma^2_w \sqbracket{\int_{\tau/\sqrt{q}}^\infty (\sqrt{q} z - \tau)^2 \, \gamma(dz)} +\sigma^2_b \\
&= 2V_{\relutau}(q) - \sigma^2_b \numberthis
\end{align*}
and, if $q^*$ is a fixed point of the above variance map, then the slope of the corresponding correlation map at $\rho = 1$ is
\begin{align*}
    \chi_{1,\st} &= \sigma^2_w \int_\R \bracket{\phi'(\sqrt{q^*} z)}^2 \, \gamma(dz) \\
    &= \sigma^2_w\bracket{\int_{-\infty}^{-\tau/\sqrt{q^*}} + \int_{\tau/\sqrt{q^*}}^\infty} \, \gamma(dz) = 2\sigma^2_w \sqbracket{1 -\Phi\bracket{\frac{\tau}{\sqrt{q^*}}}}. \numberthis
    \label{eqn:chi}
\end{align*}
As a result,
\begin{equation}
V'_{\st}(q) = 2V'_{\relutau}(q), \quad \chi_{1,\st} = 2\chi_{1,\relutau} = V'_{\st}(q^*), \quad V''_{\st}(q) = 2V''_{\relutau}(q) > 0, 
\end{equation}
and the fixed point $q^*$ is unstable when such DNN is initialized at the EoC.

\begin{figure}[t]
    \centering
    \includegraphics[width=\textwidth]{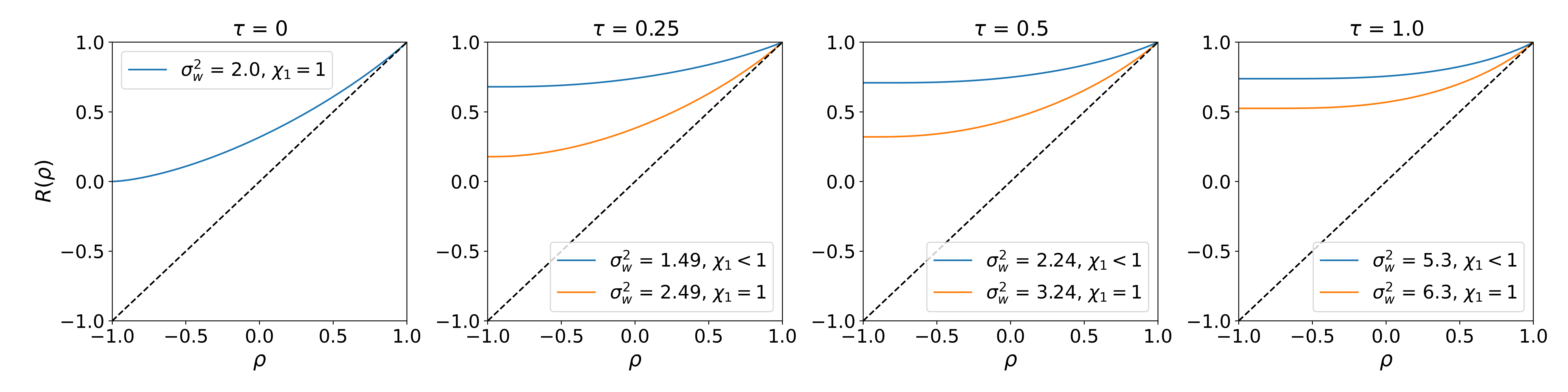}
    \caption{Correlation maps for $\relutau$ with $(\sigma_w,\sigma_b)$ on, and on either side of, the EoC, for different values of $\tau$ -- except in the case where no $q^*$ exists for those hyperparameters.}
    \label{fig:sfatrelu corr map}
\end{figure}

\begin{figure}[t]
    \centering
    \includegraphics[width=\textwidth]{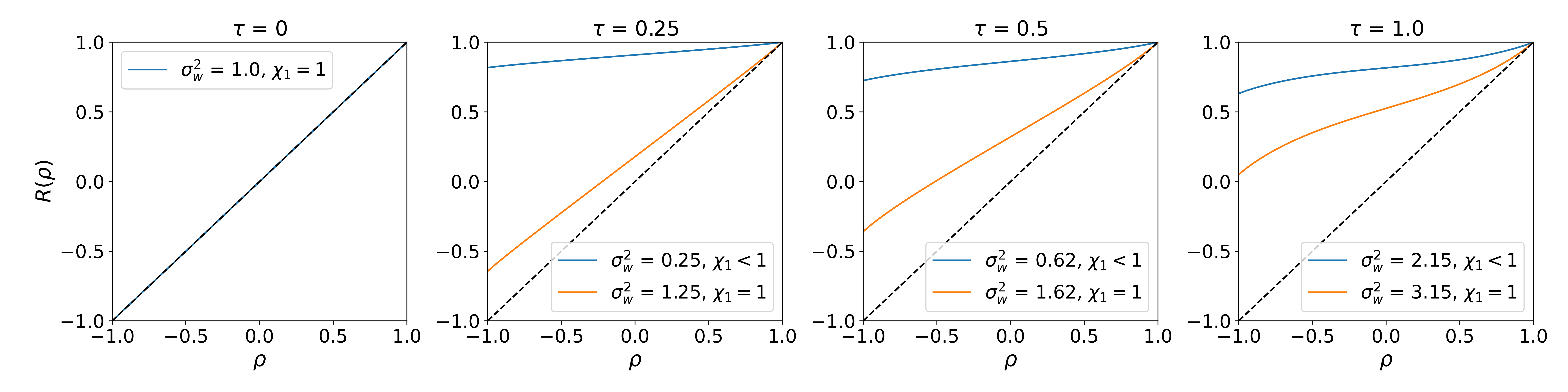}
    \caption{Correlation maps for $\st$ with $(\sigma_w,\sigma_b)$ on, and on either side of, the EoC, for different values of $\tau$ -- except in the case where no $q^*$ exists for those hyperparameters.}
    \label{fig:st corr map}
\end{figure}

Correlation maps $R_\phi(\rho)$ for $\relutau$ and $\st$ are plotted in Figures \ref{fig:sfatrelu corr map} and \ref{fig:st corr map} for different choices of the hyperparameters $(\sigma_w, \sigma_b)$, excluding those cases when $q^*>0$ does not exist. Attempting to regain the stability of $V(q)$ by setting $\chi_{1,\relutau}<1$ would result in the DNN being unstable to small perturbations, i.e. $R(\rho)$ being unstable at $\rho=1$.  Alternatively, having $\chi_{1,\relutau}\ge 1$ results in the Gaussian process having variance growing exponentially with depth and the associated exponential scaling of the gradient with depth.

\begin{figure}[t]
    \centering
    \includegraphics[width=\textwidth]{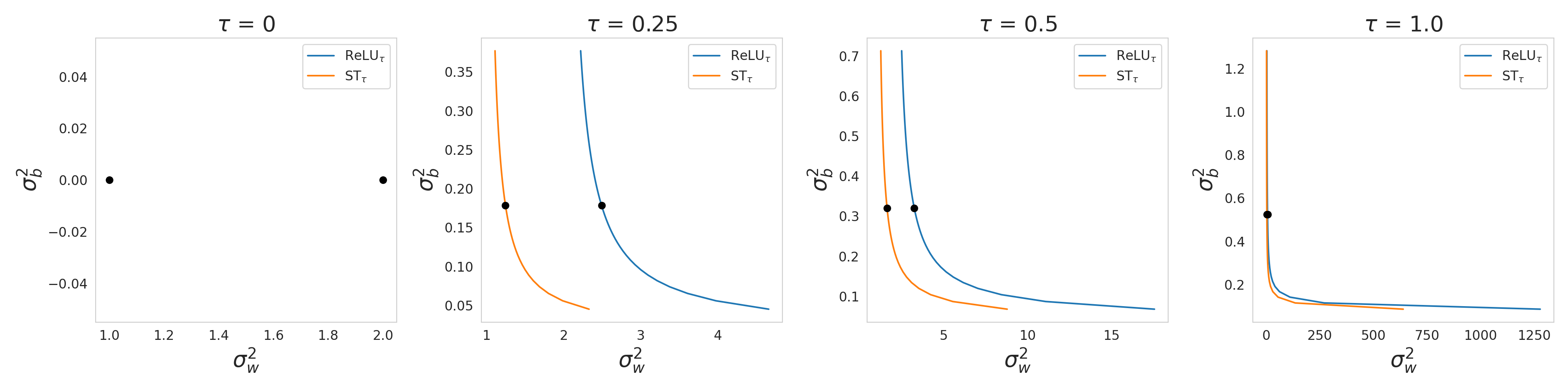}
    \caption{Edge of Chaos for $\relutau$ and $\st$ for different $\tau$. As expected, the $\sigma_w^2$ values for $\relutau$ are twice those for $\st$ for the same $q^*$. The black dot corresponds to $q^*=1$.}
    \label{fig: sfatrelu and st EoC curves}
\end{figure}

\section{Experiments demonstrating training instability for $\relutau$ and $\st$.}\label{sec: non max variants experiments}
Width 300, depth 100 feedforward networks are trained with the above activation functions to perform image classification on the MNIST dataset. The networks are initialized at the EoC using $q^*=1$, and trained with SGD with $10^{-4}$ learning rate for 200 epochs. 

When $\tau>0$ (the only exception being $\relutau$ with $s=0.5$, i.e. standard ReLU), the input is normalised to have variance $<q^*$, to give the best chance of trainability given the instability of $q^*$ in these cases (in particular the input variance is set to be $0.75$). As seen numerically however, in particular at higher sparsities, these measures are insufficient to enable trainability of $\relutau$ and $\st$ networks.

\begin{figure}[h]
    \centering
     \begin{subfigure}[b]{0.48\textwidth}
         \centering
         \includegraphics[width=\textwidth]{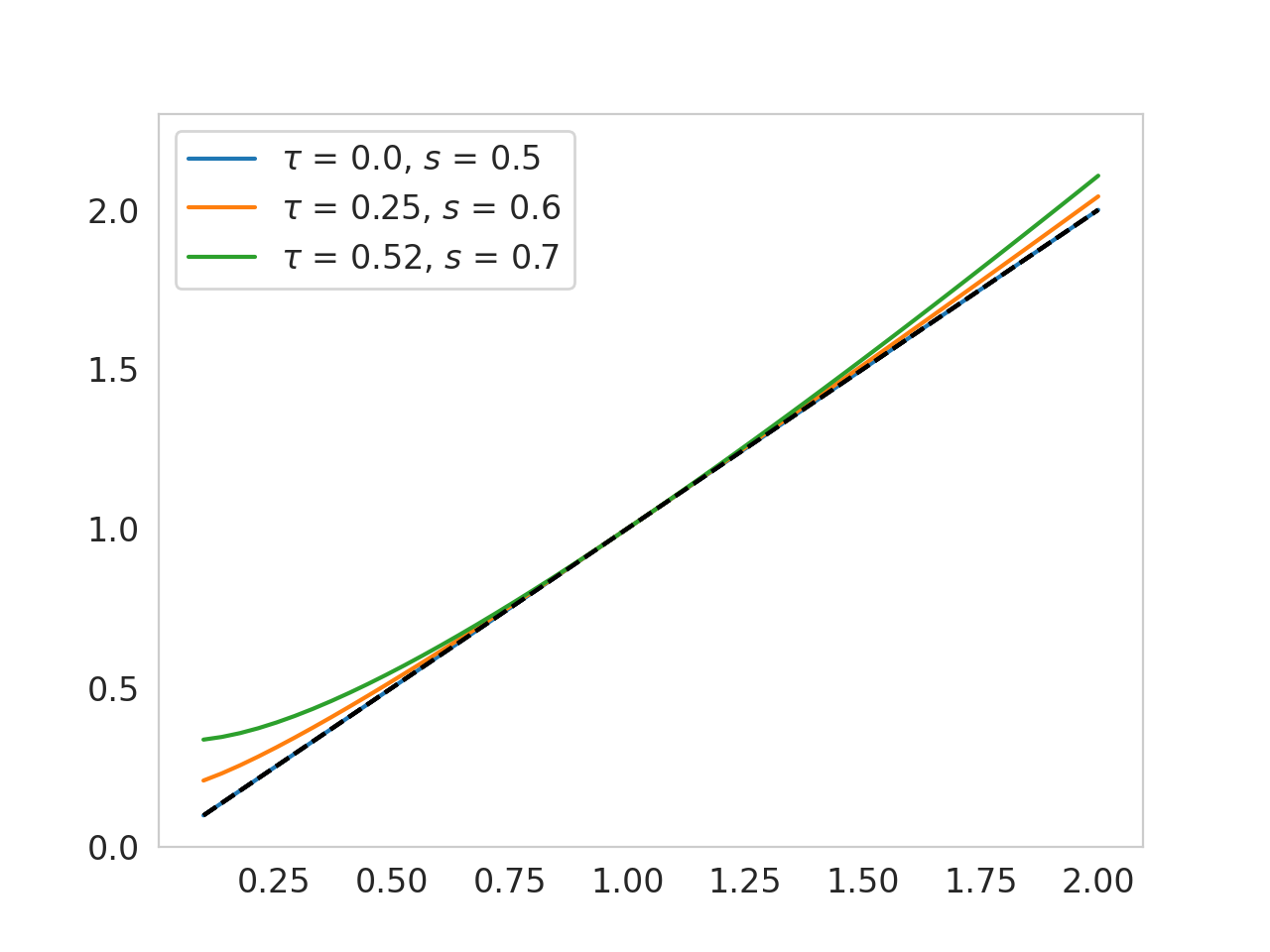}
     \end{subfigure}
     \hfill
    \begin{subfigure}[b]{0.48\textwidth}
         \centering
    \includegraphics[width=\textwidth]{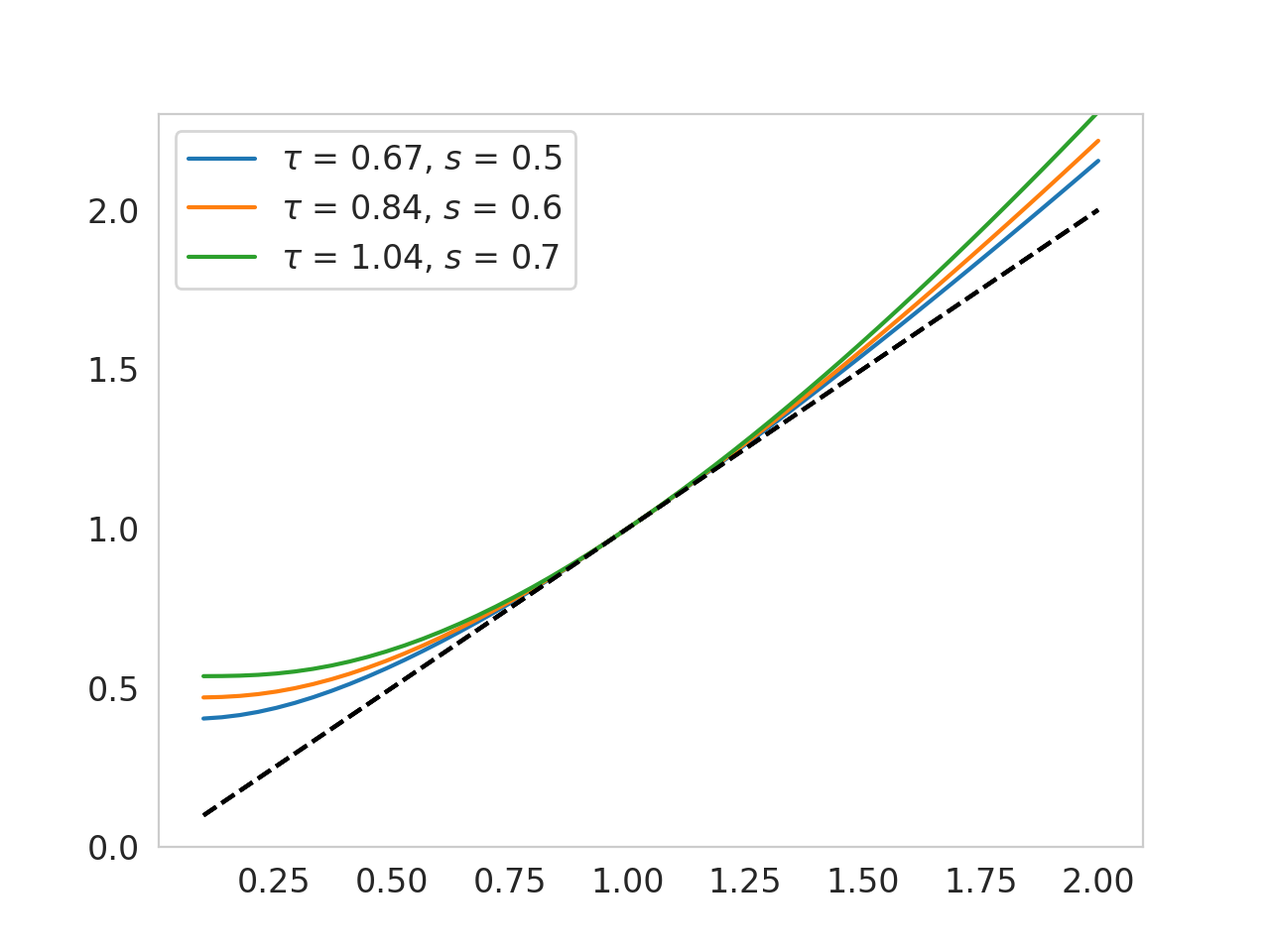}
     \end{subfigure}

\caption{Variance maps for $\relutau$ (left) and $\st$ (right) with $(\sigma_w,\sigma_b)$ at the EoC for different values of sparsity $s$.}
\label{fig:variance map sparsity}
     
\end{figure}

A variety of starting sparsity levels ([50\%, 60\%, 70\%]) are tested for each activation function, and track both their test accuracy, as well as the sparsity of their activations during evaluation, to check to what extent the initial sparsity levels are maintained during training. We plot the Variance maps $V_\phi$ for each activation function for these settings in Figure \ref{fig:variance map sparsity}. For each activation function-hyperparameter combination, we train 5 runs. Table \ref{tab: results sfatrelu_max and st_max} shows the mean and standard deviation of the test accuracy and activation sparsity after training, measured on the test set.

The results concord with our expectations based at the EoC and variance map analyses above. First, $\st$ networks fail to train at any of the tested sparsities. This reflects the notable instability of $q^*$ for $\st$ with the corresponding values of $\tau$, as shown in Figure \ref{fig:variance map sparsity}. Similarly, $\relutau$ networks also fail to train once sparsity (and thus $\tau$) grow sufficiently large.

The CNN experiments on the CIFAR10 dataset use networks with 50 layers, and 300 channels per layer, and were trained for 200 epochs with SGD and learning rate 1e-3. A single experiment was run per (activation function, sparsity) pair. Again, both activation functions fail to train at high sparsities.

\section{Additional Experimental Implementation Details}\label{sec:additional experimental details}

For both MNIST and CIFAR10, 10\% of the training set was held out as the validation set. Results reported are computed on the test set.

All layers are initialized on the EOC, with a slight modification for the first layer. In our experiments, the first layer is initialized so as to preserve the input variance in the first layer's pre-activations. This was an additional effort to try our best to protect against the instability of $q^*$ for the non-clipped activation functions. The EoC initialization is based on the assumption that the input to the layer are `post-activations', that is, the incoming vector is a Gaussian which has been passed through the activation function. However, as the input to the first layer has not been passed through an activation function -- and thus is dense, not sparse -- multiplication of the inputs by a weight matrix initialized at the EoC would substantially grow the variance, possibly to a value larger than $q^*$, which would then prevent convergence to $q^*$ with these activation functions.

Experiments were run on a single V100 GPU, and were implemented using Pytorch Lightning.

\section{Variance and correlation maps for $\crelu$ and $\cst$} \label{sec: clipped plots}

The variance map of $\crelu$ is computed as followed:
\begin{align*}
V_{\crelu}(q) &= \sigma^2_w \sqbracket{\int_{\tau/\sqrt{q}}^{(\tau+m)/\sqrt{q}} (\sqrt{q}\, z - \tau)^2 \,\gamma(dz) + \int_{(\tau+m)/\sqrt{q}}^\infty m^2 \gamma(dz)} + \sigma^2_b \\
&= \sigma^2_w(v_1(q) - v_2(q) + v_3(q)) + \sigma^2_b, \numberthis
\end{align*}
where
\begin{align*}
v_1(q) &= \int_{\tau/\sqrt{q}}^\infty (\sqrt{q}\,z - \tau)^2 \, \gamma(dz) = \int_{\tau/\sqrt{q}}^\infty (qz^2 - 2\sqrt{q}\,\tau z + \tau^2) \, \gamma(dz), \\
v_2(q) &= \int_{(\tau+m)/\sqrt{q}}^\infty (\sqrt{q}\,z - \tau)^2 \, \gamma(dz) = \int_{(\tau+m)/\sqrt{q}}^\infty (qz^2 - 2\sqrt{q}\,\tau z + \tau^2) \, \gamma(dz), \\
v_3(q) &= \int_{(\tau+m)/\sqrt{q}}^\infty m^2. \, \gamma(dz). \numberthis
\end{align*}
Notice that $v_1(q) = v_{11}(q) - \tau v_{12}(q) + \tau^2 v_{13}(q)$, where
\begin{equation}
v_{11}(q) = q \int_{\tau/\sqrt{q}}^\infty z^2 \, \gamma(dz), \quad v_{12}(q) = 2\sqrt{q} \int_{\tau/\sqrt{q}}^\infty z \, \gamma(dz), \quad v_{13}(q) = \int_{\tau/\sqrt{q}}^\infty  \gamma(dz).
\end{equation}
Therefore,
\begin{align*}
\frac{dv_{11}}{dq} &= \frac{d}{dq}\sqbracket{q \int_{\tau/\sqrt{q}}^\infty z^2 \, \gamma(dz)} \\
&= \int_{\tau/\sqrt{q}}^\infty z^2 \, \gamma(dz) + q \frac{d}{dq}\sqbracket{\int_{\tau/\sqrt{q}}^\infty z^2 \, \gamma(dz)} \\
&= \frac{\tau}{\sqrt{2\pi q}} \exp\bracket{-\frac{\tau^2}{2q}} + 1 - \Phi\bracket{\frac{\tau}{\sqrt{q}}} + \frac{\tau^3}{2\sqrt{2\pi} \, q^{3/2}} \exp\bracket{-\frac{\tau^2}{2q}}, \\
\frac{dv_{12}}{dq} &= \frac{d}{dq}\sqbracket{2\sqrt{q} \int_{\tau/\sqrt{q}}^\infty z \, \gamma(dz)} \\
&= \frac{1}{\sqrt{q}} \int_{\tau/\sqrt{q}}^\infty z \, \gamma(dz) + 2\sqrt{q} \frac{d}{dq} \sqbracket{\int_{\tau/\sqrt{q}}^\infty z \, \gamma(dz)} \\
&= \frac{1}{\sqrt{2\pi q}}  \exp\bracket{-\frac{\tau^2}{2q}} + \frac{\tau^2}{\sqrt{2\pi} \, q^{3/2}} \exp\bracket{-\frac{\tau^2}{2q}}, \\
\frac{dv_{13}}{dq} &= \frac{d}{dq}\sqbracket{\int_{\tau/\sqrt{q}}^\infty  \gamma(dz)} = \frac{\tau}{2\sqrt{2\pi} \, q^{3/2}} \exp\bracket{-\frac{\tau^2}{2q}}.
\end{align*}
Hence
\begin{align*}
\frac{dv_1}{dq} 
&= \frac{dv_{11}}{dq} - \tau \frac{dv_{12}}{dq} + \tau^2 \frac{dv_{13}}{dq} \\
&= \frac{\tau}{\sqrt{2\pi q}} \exp\bracket{-\frac{\tau^2}{2q}} + 1 - \Phi\bracket{\frac{\tau}{\sqrt{q}}} + \frac{\tau^3}{2\sqrt{2\pi} \, q^{3/2}} \exp\bracket{-\frac{\tau^2}{2q}} \\
&\phantom{=}- \frac{\tau}{\sqrt{2\pi q}}  \exp\bracket{-\frac{\tau^2}{2q}} - \frac{\tau^3}{\sqrt{2\pi} \, q^{3/2}} \exp\bracket{-\frac{\tau^2}{2q}} + \frac{\tau^3}{2\sqrt{2\pi} \, q^{3/2}} \exp\bracket{-\frac{\tau^2}{2q}} \\
&= 1 - \Phi\bracket{\frac{\tau}{\sqrt{q}}}. \numberthis
\end{align*}
One could similarly derive
\begin{align*}
\frac{dv_2}{dq} &= \frac{\tau+m}{\sqrt{2\pi q}} \exp\bracket{-\frac{(\tau+m)^2}{2q}} + 1 - \Phi\bracket{\frac{\tau+m}{\sqrt{q}}} + \frac{(\tau+m)^3}{2\sqrt{2\pi} \, q^{3/2}} \exp\bracket{-\frac{(\tau+m)^2}{2q}} \\
&\phantom{=}- \frac{\tau}{\sqrt{2\pi q}}  \exp\bracket{-\frac{(\tau+m)^2}{2q}} - \frac{\tau(\tau+m)^2}{\sqrt{2\pi} \, q^{3/2}} \exp\bracket{-\frac{(\tau+m)^2}{2q}} \\
&\phantom{=}+ \frac{\tau^2 (\tau+m)}{2\sqrt{2\pi} \, q^{3/2}} \exp\bracket{-\frac{(\tau+m)^2}{2q}} \\
&= \frac{m}{\sqrt{2\pi q}} \exp\bracket{-\frac{(\tau+m)^2}{2q}} + 1 - \Phi\bracket{\frac{\tau+m}{\sqrt{q}}} + \frac{m^2 (\tau+m)}{2\sqrt{2\pi} \, q^{3/2}} \exp\bracket{-\frac{(\tau+m)^2}{2q}}
\numberthis
\end{align*}
Finally,
\begin{equation}
\frac{dv_3}{dq} = \frac{m^2(\tau+m)}{2\sqrt{2\pi}\, q^{3/2}} \exp\bracket{-\frac{(\tau+m)^2}{2q}},
\end{equation}
so 
\begin{align*}
\frac{dV_{\crelu}}{dq} &= \sigma^2_w \sqbracket{\frac{dv_1}{dq} - \frac{dv_2}{dq} + \frac{dv_3}{dq}} \\
&= \sigma^2_w \sqbracket{\Phi\bracket{\frac{\tau+m}{\sqrt{q}}} - \Phi\bracket{\frac{\tau}{\sqrt{q}}} - \frac{m}{\sqrt{2\pi q}} \exp\bracket{-\frac{(\tau+m)^2}{2q}}} \\
&= \sigma^2_w \sqbracket{\frac{1}{2}\text{erf}\bracket{\frac{\tau+m}{\sqrt{2q}}} - \frac{1}{2}\text{erf}\bracket{\frac{\tau}{\sqrt{2q}}} - \frac{m}{\sqrt{2\pi q}} \exp\bracket{-\frac{(\tau+m)^2}{2q}}}. \numberthis
\end{align*}
The slope of the associated correlation map at $\rho = 1$ for DNNs initialized at the fixed point $q^*$ is
\begin{align*}
\chi_{1,\crelu} = \sigma^2_w \int_{\tau/\sqrt{q^*}}^{(\tau+m)/\sqrt{q^*}} \gamma(dz) &= \sigma^2_w \bracket{\Phi\bracket{\frac{\tau+m}{\sqrt{q^*}}} - \Phi\bracket{\frac{\tau}{\sqrt{q^*}}}} \\
&= \sigma^2_w \sqbracket{\frac{1}{2}\text{erf}\bracket{\frac{\tau+m}{\sqrt{2q^*}}} - \frac{1}{2}\text{erf}\bracket{\frac{\tau}{\sqrt{2q^*}}}}. \numberthis
\end{align*}
Therefore
\begin{equation}
\chi_{1,\crelu} = V'_{\crelu}(q^*) + \frac{\sigma^2_w m}{\sqrt{2\pi q^*}} \exp\bracket{-\frac{(m+\tau)^2}{2q^*}} > V'_{\crelu}(q^*),
\end{equation}
which makes any fixed points $q^*$ of $V_{\crelu}$ at the EoC locally stable, as described in Sec.\ \ref{sec:clipped}.

For the other activation functions $\cst$, the variance map is:
\begin{align*}
V_{\cst}(q) &= \sigma^2_w \bigg[\int_{-\infty}^{-(\tau+m)/\sqrt{q}} m^2 \, \gamma(dz) + \int_{-(\tau+m)/\sqrt{q}}^{-\tau/\sqrt{q}} (\sqrt{q} z + \tau)^2 \, \gamma(dz) \\
&\phantom{=}+ \int_{\tau/\sqrt{q}}^{(\tau+m)/\sqrt{q}} (\sqrt{q}z - \tau)^2 \, \gamma(dz) + \int_{(\tau+m)/\sqrt{q}}^\infty m^2 \, \gamma(dz) \bigg] + \sigma^2_b \\
&= 2\sigma^2_w \sqbracket{\int_{\tau/\sqrt{q}}^{(\tau+m)/\sqrt{q}} (\sqrt{q}z - \tau)^2 \, \gamma(dz) + \int_{(\tau+m)/\sqrt{q}}^\infty m^2 \, \gamma(dz)} +\sigma^2_b \\
&= 2V_{\crelu}(q) - \sigma^2_b, \numberthis
\end{align*}
and the slope of the associated correlation map is
\begin{align*}
\chi_{1,\cst} &= \sigma^2_w \bracket{\int_{-(\tau+m)/\sqrt{q^*}}^{-\tau/\sqrt{q^*}} + \int_{\tau/\sqrt{q^*}}^{(\tau+m)/\sqrt{q^*}}} \gamma(dz) \\ &= 2\sigma^2_w \int_{\tau/\sqrt{q^*}}^{(\tau+m)/\sqrt{q^*}} \gamma(dz) = 2\chi_{1,\crelu}. \numberthis
\end{align*}
This yields $V'_{\cst}(q^*) = 2V'_{\crelu}(q^*)$, and that
\begin{equation}
\chi_{1,\cst} = 2\chi_{1,\crelu} = V'_{\cst}(q^*) + \frac{\sqrt{2} \, \sigma^2_w m}{\sqrt{\pi q^*}} \exp\bracket{-\frac{(m+\tau)^2}{2q^*}} > V'_{\cst}(q^*),
\end{equation}
which shows that any fixed points $q^*$ of $V_{\cst}$ at the EoC are locally stable as well.

\begin{figure}[h]
    \centering\includegraphics[width=\textwidth]{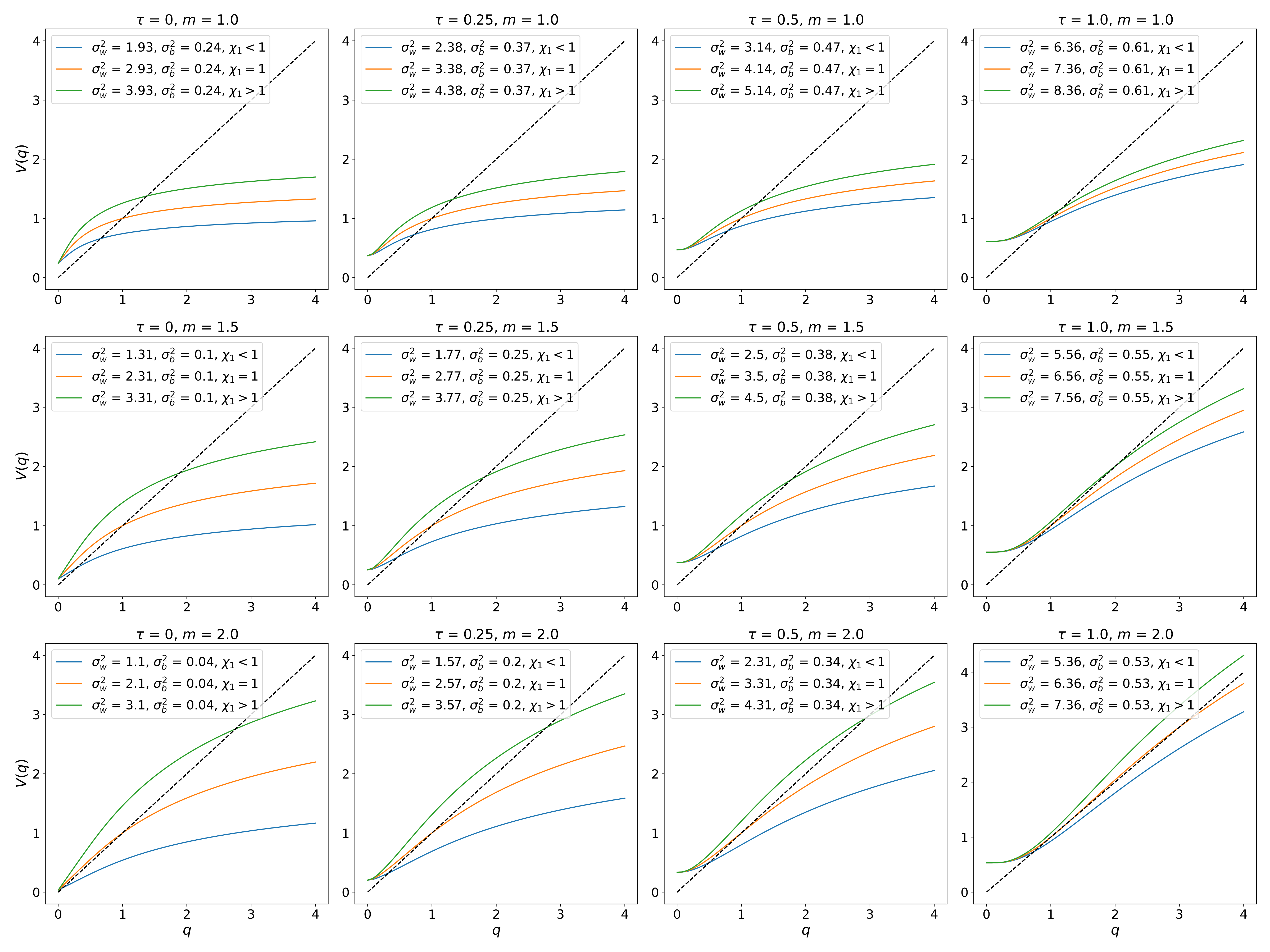}
    \caption{Variance maps for $\crelu$ with $(\sigma_w,\sigma_b)$ at and on either side of the EoC, for different $s$ and $m$, with $q^* = 1$. From top to bottom, the rows correspond to $m=1$, $m=1.5$, and $m=2$, and from left to right the columns correspond to $\tau=0$, $\tau=0.25$, $\tau=0.5$ and $\tau=1$.}
    \label{fig: sfatrelu max variance map}
\end{figure}

\begin{figure}[h]
    \centering
    \includegraphics[width=\textwidth]{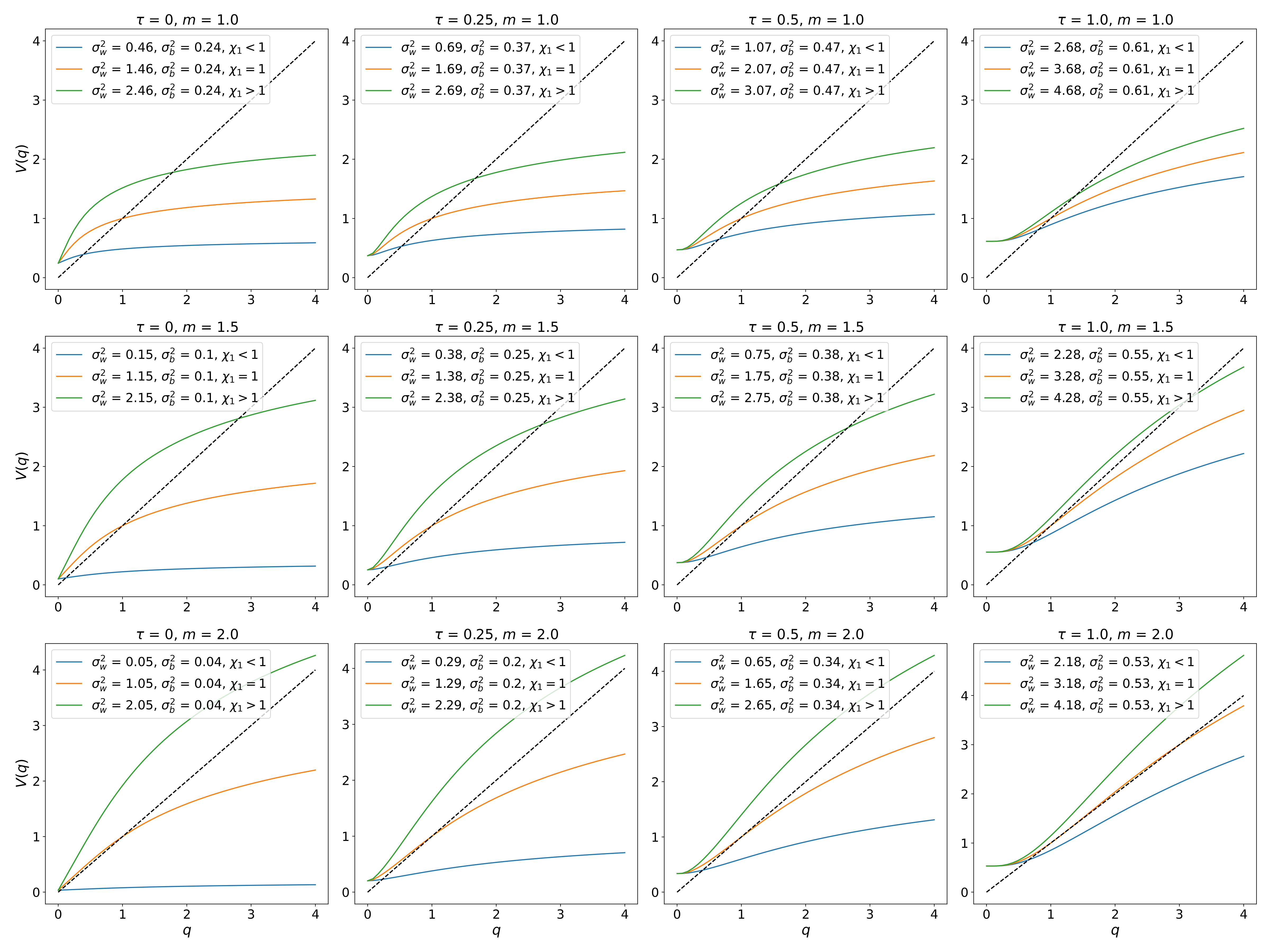}
    \caption{Variance maps for $\cst$ with $(\sigma_w,\sigma_b)$ at and on either side of the EoC, for different $\tau$ and $m$, with $q^* = 1$. From top to bottom, the rows correspond to $m=1$, $m=1.5$, and $m=2$, and from left to right the columns correspond to $\tau=0$, $\tau=0.25$, $\tau=0.5$ and $\tau=1$.}
    \label{fig: st max variance map}
\end{figure}

\begin{figure}[h]
    \centering
    \includegraphics[width=\textwidth]{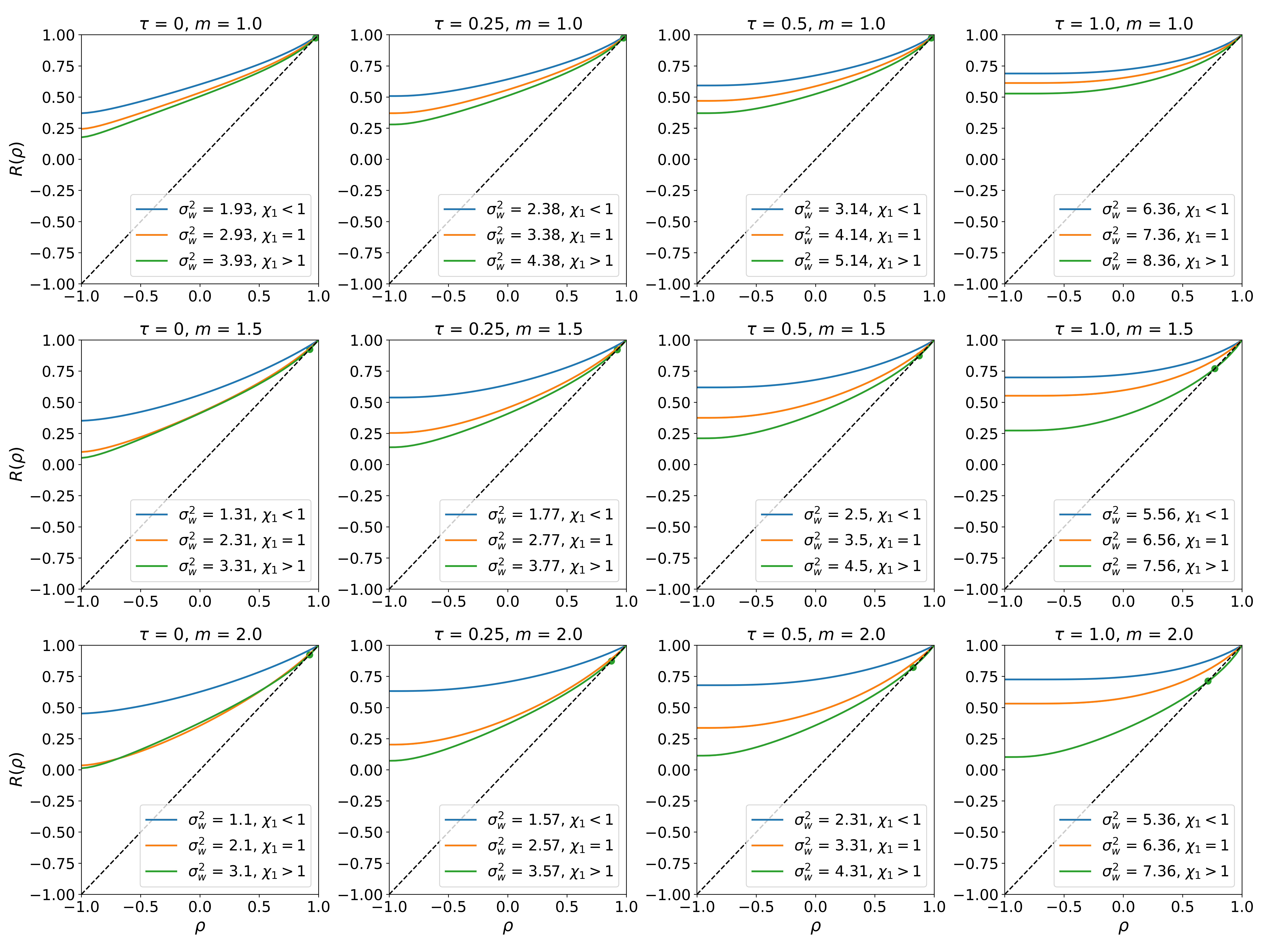}
    \caption{Correlation maps for $\crelu$ with $(\sigma_w,\sigma_b)$ at and on either side of the EoC, for different $\tau$ and $m$. From top to bottom, the rows correspond to $m=1$, $m=1.5$, and $m=2$, and from left to right the columns correspond to $\tau=0$, $\tau=0.25$, $\tau=0.5$ and $\tau=1$. Fixed points $\rho^*<1$ (where the correlation map crosses the diagonal $R(\rho) = \rho$) are indicated by round markers on the relevant curves.}
    \label{fig: sfatrelu max corr map}
\end{figure}

\begin{figure}[h]
    \centering
    \includegraphics[width=\textwidth]{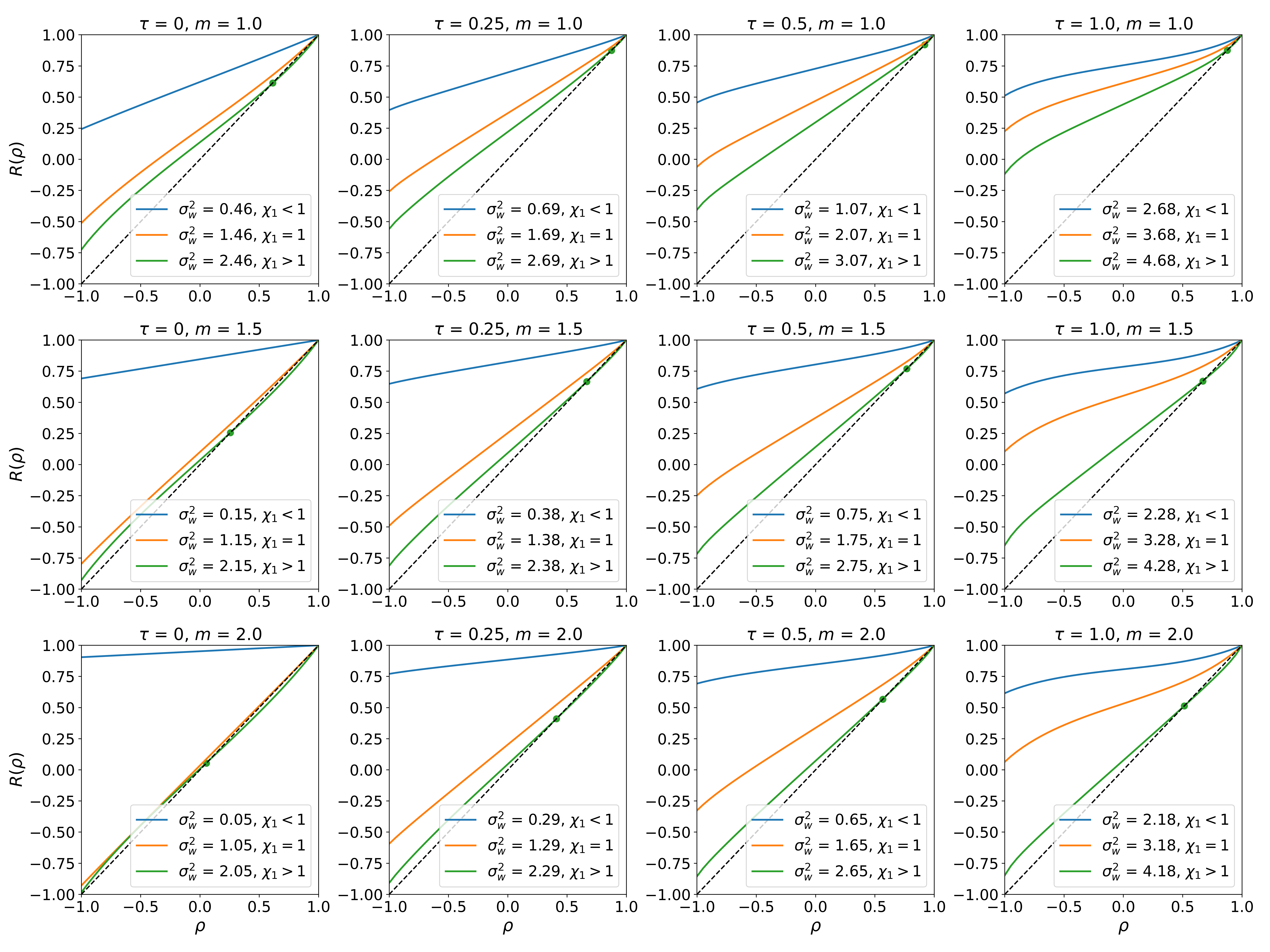}
    \caption{Correlation maps for $\cst$ with $(\sigma_w,\sigma_b)$ at and on either side of the EoC, for different $\tau$ and $m$. From top to bottom, the rows correspond to $m=1$, $m=1.5$, and $m=2$, and from left to right the columns correspond to $\tau=0$, $\tau=0.25$, $\tau=0.5$ and $\tau=1$. Fixed points $\rho^*<1$ (where the correlation map crosses the diagonal $R(\rho) = \rho$) are indicated by round markers on the relevant curves.}
    \label{fig: st max corr map}
\end{figure}

\clearpage

\section{Relationship between $\sigma_w^2$ and the expected sparsity at the EoC, control of  variance of the input-output Jacobian.} \label{sec: clipped jacobian}

Recall that the `expected sparsity' with respect to an activation function $\phi$ could be computed under the usual mean-field assumption (that each pre-activation $Z$ follows a normal distribution with variance $q^*$):
\begin{equation}
s = \mathbb{P}(\phi(Z) = 0), \quad Z \sim \mathcal{N}(0,q^*).
\end{equation}
When studying DNNs with sparsifying activation functions ($\phi = \relutau$, $\st$, $\crelu$ or  $\cst$), we would like to find the optimal $\hat{\tau}$ such that the DNNs achieve the expected sparsity $s$. The optimal $\tau$ is related to the $s$ and $q^*$ as followed:
\begin{itemize}
    \item For $\phi = \relutau$ or $\crelu$:
    \begin{align*} \label{eq: sfatrelu tau}
    \hat{\tau} = \hat{\tau}_\phi(s, q^*) &:= \inf_\tau \{\tau \,|\, \mathbb{P}(\phi(Z) = 0) \geq s, \, Z \sim \mathcal{N}(0,q^*) \} \\
    &= \inf_\tau \{\tau \,|\, \mathbb{P}(Z < \tau) \geq s, \, Z \sim \mathcal{N}(0,q^*) \} \\
    &= \sqrt{q^*} \Phi^{-1}(s). \numberthis
    \end{align*}
    \item and for $\phi = \st$ or $\cst$,
    \begin{align*}
    \hat{\tau} = \hat{\tau}_\phi(s, q^*) &:= \inf_\tau \{\tau \,|\, \mathbb{P}(\phi(Z) = 0) \geq s, \, Z \sim \mathcal{N}(0,q^*) \} \\
    &= \inf_\tau \{\tau \,|\, \mathbb{P}(-\tau < Z < \tau) \geq s, \, Z \sim \mathcal{N}(0,q^*) \} \\
    &= \sqrt{2q^*} \text{erf}^{-1}(s) = \sqrt{q^*} \Phi^{-1}\bracket{\frac{s+1}{2}}. \numberthis 
    \end{align*}
\end{itemize}
There is a simple formula to compute $\sigma^2_w$ for initialising the DNNs at the EoC ($\chi_{1,\phi} = 1$) with activation functions $\phi = \relutau$ or $\st$. In particular,
\begin{itemize}
    \item for $\phi = \relutau$:
\begin{equation}
    \sigma_w^2 = \sigma_w^2(s,q^*) := \left(1 - \Phi\left(\frac{\hat{\tau}_{\phi}(s,q^*)}{\sqrt{q^*}}\right) \right)^{-1} = \left(1 - \Phi\left(\Phi^{-1}(s)\right) \right)^{-1} = \frac{1}{1-s}, \label{eq: sw2 sfatrelu}
\end{equation}
    \item and for $\st$, using the identity $\text{erf}^{-1}(s) = \frac{1}{\sqrt{2}} \Phi^{-1}\left( \frac{s+1}{2}\right)$:
\begin{align*}
    \sigma_w^2 = \sigma_w^2(s,q^*) &:= \frac{1}{2}\left(1 - \Phi\left(\frac{\hat{\tau}(s,q^*)}{\sqrt{q^*}}\right) \right)^{-1} \\
    &= \frac{1}{2}\left(1 - \Phi\left(\Phi^{-1}\left(\frac{s+1}{2}\right)\right) \right)^{-1} = \frac{1}{1-s}. \numberthis \label{eq: sw2 soft threshold}
\end{align*}
\end{itemize}
Notice that the $\sigma^2_w$ above are independent of the variance of the pre-activation $q^*$. This is not true for the cases of the clipped activations $\phi = \crelu$ or $\cst$, instead, the initialization $\sigma^2_w$ at the EoC now depends on $q^*$, $m$, and $s$. In fact, 
\begin{itemize}
    \item for $\crelu$, using the identity $\Phi^{-1}(s) = \sqrt{2}\text{erf}^{-1}(2s-1)$:
    \begin{align*}
    \sigma_w^2 &= 2 \left( \text{erf}\left(\frac{m + \tau}{\sqrt{2q^*}}\right) - \text{erf}\left(\frac{\tau}{\sqrt{2q^*}}\right) \right)^{-1} \\
    &= 2 \left( \text{erf}\left(\frac{m + \sqrt{2q^*} \text{erf}^{-1}(2s-1)}{\sqrt{2q^*}}\right) - \text{erf}\left(\text{erf}^{-1}(2s-1)\right) \right)^{-1} \\
    &= 2 \left( \text{erf}\left(\frac{m}{\sqrt{2q^*}} + \text{erf}^{-1}(2s-1)\right) - 2s+1\right)^{-1}, \numberthis \label{eq: sw2 sfatrelu max}
    \end{align*}
    \item and for $\cst$:
    \begin{align*}
    \sigma_w^2 &= \left( \text{erf}\left(\frac{m + \tau}{\sqrt{2q^*}}\right) - \text{erf}\left(\frac{\tau}{\sqrt{2q^*}}\right) \right)^{-1} \\
    &= \left( \text{erf}\left(\frac{m + \sqrt{2q^*} \text{erf}^{-1}(s)}{\sqrt{2q^*}}\right) - \text{erf}\left(\text{erf}^{-1}(s)\right) \right)^{-1} \\
    &= \left( \text{erf}\left(\frac{m}{\sqrt{2q^*}} + \text{erf}^{-1}(s)\right) - s \right)^{-1}. \numberthis \label{eq: sw2 soft threshold max}
    \end{align*}
\end{itemize}

Despite the fact that that $\sigma_w^2$ is now dependent on $q^*$, the derivatives of $\crelu$ and $\cst$ take only values of zero or one, like those of their non-clipped counterparts, and thus the moments of the spectra as computed by \eqref{eq: mu_k} are independent of $k$, see Table \ref{tab: spectrum}. Once again, for both clipped activation functions, $\mu_k = \mu$ for all $k$ such that $\mu_1/\mu_2^2 = \sigma_w^2$ as was the case for $\relutau$ and $\st$. For both $\crelu$ ($s\geq0.5$) and $\cst$ ($s\geq0$), from Equations \eqref{eq: sw2 sfatrelu max} and \eqref{eq: sw2 soft threshold max} that
\begin{align}
\sigma_w^2 \longrightarrow \frac{1}{1-s} \qquad \text{as} \qquad q^* \longrightarrow 0,   
\end{align}
meaning that $\sigma_{JJ^\top}^2$ grows linearly with $L$ with these activation functions, with at least the same rate as their unclipped counterparts.

Nonetheless, the prospect of a stable EoC initialization stands as a substantial improvement over the unclipped variants, and as we show in Section \ref{sec: max variant experiments}, this appears sufficient to enable trainability of very deep networks with sparse activations in practice.

\begin{table}[h]
\centering
\caption{Summary statistics for the Jacobian's spectrum}
\label{tab: spectrum}
\begin{tabular}{lcccc}
\hline
& $\mu_k \ (= \mu)$       & $\sigma_w^2$  & $M_{D^2}(z)$ & $\sigma_{JJ^\top}^2$ \\ \cline{2-5} 
$\relutau$ & $\displaystyle{1 - \Phi\left(\frac{\tau}{\sqrt{q^*}}\right)}$ &  $\displaystyle{\frac{1}{\mu}}$  &  $\displaystyle{\frac{1}{\sigma_w^2} \frac{1}{1-z}}$  & $L\left(\sigma_w^2 - 1 - s_1 \right)$ \\
$\st$ & $\DS{2\bracket{1 - \Phi\left(-\frac{\tau}{\sqrt{q^*}} \right)}}$ &   $\DS{\frac{1}{\mu}}$ & $\DS{\frac{1}{\sigma_w^2} \frac{1}{1-z}}$  & $L\left(\sigma_w^2 - 1 - s_1 \right)$\\
$\crelu$ & $\DS{\frac{1}{2} \left( \text{erf}\left(\frac{m + \tau}{\sqrt{2q^*}}\right) - \text{erf}\left(\frac{\tau}{\sqrt{2q^*}}\right) \right)}$ & $\DS{\frac{1}{\mu}}$ & $\DS{\frac{1}{\sigma_w^2} \frac{1}{1-z}}$ & $L\left(\sigma_w^2 - 1 - s_1 \right)$\\
$\cst$ & $\DS{\text{erf}\left(\frac{m + \tau}{\sqrt{2q^*}}\right) - \text{erf}\left(\frac{\tau}{\sqrt{2q^*}}\right)}$ & $\DS{\frac{1}{\mu}}$  & $\frac{1}{\sigma_w^2} \frac{1}{1-z}$ & $L\left(\sigma_w^2 - 1 - s_1 \right)$ \\ \hline
\end{tabular}
\end{table}

\clearpage

\clearpage

\section{Failure to train for high sparsity and low magnitude clipping}
\label{sec: loss function}

\begin{figure}[h]
    \centering
     \begin{subfigure}[b]{0.48\textwidth}
         \centering
         \includegraphics[width=\textwidth]{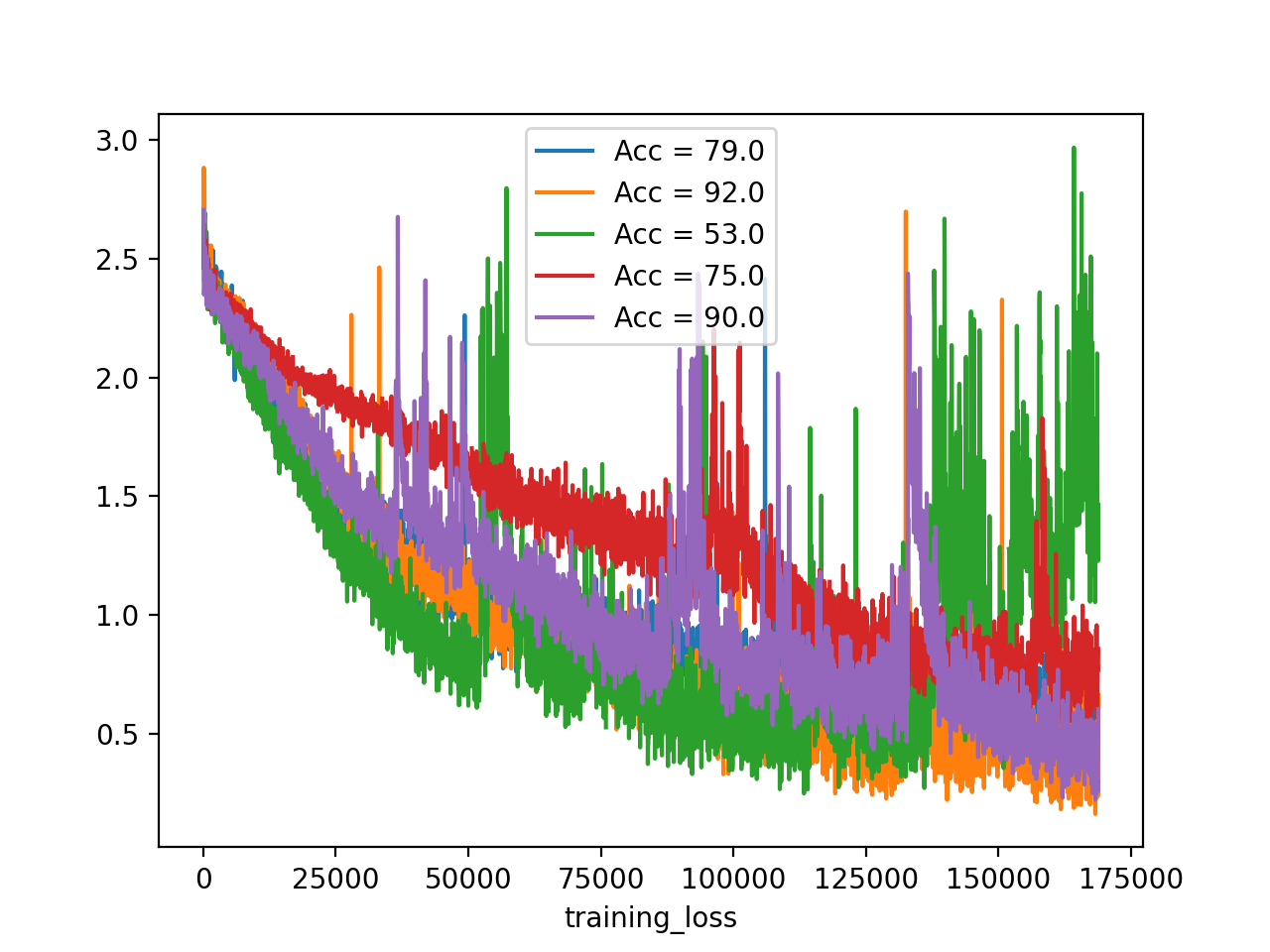}
         \caption{$\cst$, $s=0.85$, $m=0.81$}
        \label{fig:sfatrelu train curve}
     \end{subfigure}
     \hfill
    \begin{subfigure}[b]{0.48\textwidth}
         \centering
    \includegraphics[width=\textwidth]{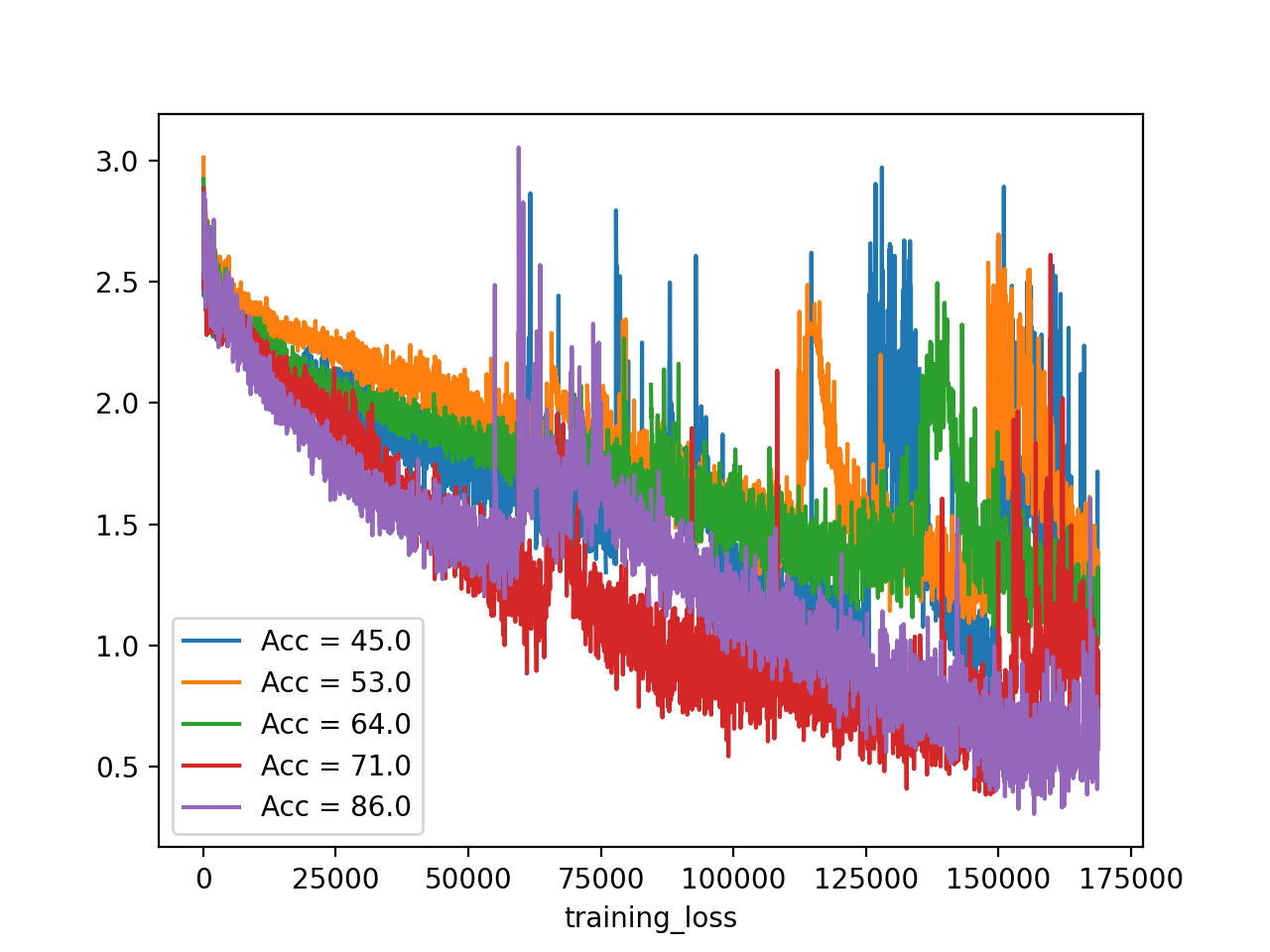}
    \caption{$\cst$, $s=0.9$, $m=0.76$}
    \label{fig:st train curve}
     \end{subfigure}
     \caption{The loss curves during training for each of the 5 runs, for two activation function and hyperparameter configurations with high activation sparsity which achieve low test accuracy as a result of slow and/or unstable training. The legend shows the corresponding final test accuracy achieved by each model.}
     \label{fig: train loss}
\end{figure}

The second failure mode discussed in Section \ref{sec: max variant experiments} appears not to be a failure of the initialization, and thus rather suggests the existence of challenges when training networks with such a high level of activation sparsity combined with small values of $m$. Indeed, inspecting the training loss curves (see e.g.\ Figure \ref{fig: train loss}) suggests that though training begins in an expected fashion, the problem for those runs exhibiting low accuracy is a combination of slow training, and instability later on in training---in these cases, we see the emergence of large spikes in training loss, which (to varying degrees) are not fully recovered from. Investigating the source and/or dynamics of the observed training speed and stability issues late in training is beyond the scope of this paper, and we defer it to future work.

\section{Experiments with shallower networks}

We have repeat the DNN experiments on MNIST from the main paper, but with depth 30 instead of depth 100. The results are shown in Tables \ref{tab: shallow dnns sfatrelu and st} and \ref{tab: shallow dnns sfatrelu_max and st_max}

We can see from these new results that the key conclusions from the 100 layer experiments hold true with 30 layers too. While $\relutau$ and $\st$ networks fail to train consistently once sparsity reaches 80\% and 70\% respectively, $\crelu$ and $\cst$ networks can train consistently to high accuracy with activation sparsity of 90\%.

\begin{table}[]
    \centering
\begin{tabular}{lllcccc}
\toprule
               &      &      & \multicolumn{2}{l}{Test Accuracy} & \multicolumn{2}{l}{Test Sparsity} \\
               &      &      &          mean &   std &                  mean &   std \\
 & $s$ & $\tau$ &               &       &                       &       \\
\midrule
$\relutau$ & 0.70 & 0.52 &          0.93 &  0.01 &                  0.67 &  0.01 \\
               & 0.80 & 0.84 &          0.60 &  0.46 &                  0.47 &  0.42 \\
               & 0.85 & 1.04 &          0.27 &  0.37 &                  0.18 &  0.38 \\
               & 0.90 & 1.28 &          0.10 &  0.00 &                  0.02 &  0.00 \\
\midrule
$\st$ & 0.70 & 1.04 &          0.27 &  0.38 &                  0.14 &  0.31 \\
               & 0.80 & 1.28 &          0.10 &  0.00 &                  0.00 &  0.00 \\
               & 0.85 & 1.44 &          0.10 &  0.00 &                  0.01 &  0.00 \\
               & 0.90 & 1.64 &          0.27 &  0.37 &                  0.19 &  0.38 \\
\bottomrule
\end{tabular}
    \caption{Experimental results for $\relutau$ and $\st$ DNNs with 30 layers on MNIST, with sparsity $s$ up to $0.9$. The mean and standard deviation of 5 runs are reported for both test accuracy and average activation sparsity calculated on the test set.}
    \label{tab: shallow dnns sfatrelu and st}
\end{table}

\begin{table}[]
    \centering
    \begin{tabular}{llllllcccc}
\toprule
                   &      &      &      &     &       & \multicolumn{2}{l}{Test Accuracy} & \multicolumn{2}{l}{Test Sparsity} \\
                   &      &      &      &     &       &          mean &   std &                  mean &   std \\
 & $s$ & $\tau$ & $m$ & $V'(q^*)$ & $V''(q^*)$ &               &       &                       &       \\
\midrule
$\crelu$ & 0.70 & 0.52 & 1.05 & 0.5 & -0.37 &          0.91 &  0.01 &                  0.70 &  0.01 \\
                   &      &      & 1.45 & 0.7 & -0.31 &          0.92 &  0.00 &                  0.70 &  0.01 \\
                   &      &      & 2.05 & 0.9 & -0.04 &          0.93 &  0.01 &                  0.69 &  0.01 \\
                   & 0.80 & 0.84 & 0.89 & 0.5 & -0.24 &          0.90 &  0.02 &                  0.80 &  0.01 \\
                   &      &      & 1.27 & 0.7 & -0.12 &          0.92 &  0.01 &                  0.80 &  0.01 \\
                   &      &      & 1.85 & 0.9 &  0.21 &          0.92 &  0.01 &                  0.78 &  0.02 \\
                   & 0.85 & 1.04 & 0.81 & 0.5 & -0.14 &          0.90 &  0.01 &                  0.85 &  0.01 \\
                   &      &      & 1.17 & 0.7 &  0.02 &          0.91 &  0.01 &                  0.84 &  0.01 \\
                   &      &      & 1.74 & 0.9 &  0.41 &          0.92 &  0.01 &                  0.84 &  0.01 \\
                   & 0.90 & 1.28 & 0.72 & 0.5 &  0.00 &          0.85 &  0.06 &                  0.90 &  0.01 \\
                   &      &      & 1.06 & 0.7 &  0.23 &          0.90 &  0.03 &                  0.89 &  0.01 \\
                   &      &      & 1.61 & 0.9 &  0.69 &          0.91 &  0.01 &                  0.74 &  0.02 \\
\midrule
$\cst$ & 0.70 & 1.04 & 0.81 & 0.5 & -0.14 &          0.91 &  0.01 &                  0.70 &  0.00 \\
                   &      &      & 1.17 & 0.7 &  0.02 &          0.92 &  0.01 &                  0.69 &  0.01 \\
                   &      &      & 1.74 & 0.9 &  0.41 &          0.93 &  0.01 &                  0.66 &  0.02 \\
                   & 0.80 & 1.28 & 0.72 & 0.5 &  0.00 &          0.90 &  0.01 &                  0.80 &  0.01 \\
                   &      &      & 1.06 & 0.7 &  0.23 &          0.91 &  0.01 &                  0.79 &  0.01 \\
                   &      &      & 1.61 & 0.9 &  0.69 &          0.92 &  0.01 &                  0.49 &  0.16 \\
                   & 0.85 & 1.44 & 0.67 & 0.5 &  0.11 &          0.88 &  0.02 &                  0.84 &  0.01 \\
                   &      &      & 1.00 & 0.7 &  0.39 &          0.91 &  0.01 &                  0.84 &  0.01 \\
                   &      &      & 1.53 & 0.9 &  0.89 &          0.27 &  0.37 &                  0.44 &  0.21 \\
                   & 0.90 & 1.64 & 0.62 & 0.5 &  0.28 &          0.77 &  0.17 &                  0.90 &  0.01 \\
                   &      &      & 0.93 & 0.7 &  0.63 &          0.91 &  0.01 &                  0.89 &  0.01 \\
                   &      &      & 1.44 & 0.9 &  1.20 &          0.11 &  0.00 &                  0.25 &  0.01 \\
\bottomrule
\end{tabular}

    \caption{Experimental results for $\crelu$ and $\cst$ DNNs wth 30 layers on MNIST, with sparsity $s$ up to $0.9$. The mean and standard deviation of 5 runs are reported for both test accuracy and average activation sparsity calculated on the test set.}
    \label{tab: shallow dnns sfatrelu_max and st_max}
\end{table}

\end{document}